\PassOptionsToPackage{table}{xcolor}
\documentclass[11pt]{article}
\usepackage{pifont}
\usepackage[final]{acl}
\usepackage{times}
\usepackage{latexsym}
\usepackage[T1]{fontenc}
\usepackage[utf8]{inputenc}
\usepackage{microtype}
\usepackage{inconsolata}
\usepackage{graphicx}
\graphicspath{{./}{../}}
\usepackage{tabularx}
\usepackage{booktabs}
\usepackage[table]{xcolor}
\usepackage{multirow}
\usepackage{hyperref}
\usepackage{listings}
\usepackage{amsmath}
\usepackage{amssymb}
\usepackage{algorithm}
\usepackage{algpseudocode}
\usepackage{makecell}

\newcommand{\cmark}{\textcolor{green!60!black}{\ding{51}}}
\newcommand{\xmark}{\textcolor{red}{\ding{55}}}

\title{YouRA: A Persistent-State Architecture for Evidence-Traceable Autonomous Research Agents}
\author{
Yoonkyu Woo$^{1,2}$\thanks{Co-first Authors} \quad
Woojin Lee$^{1}$\footnotemark[1] \quad
Jin-xia Huang$^{1,2}$\thanks{Corresponding Author} \\
\\
$^1$Electronics and Telecommunications Research Institute, Republic of Korea \\
$^2$Department of Artificial Intelligence, University of Science and Technology, Republic of Korea \\
\texttt{\{yoon303, writerwoody, hgh\}@etri.re.kr} \\
}

\begin{document}
\maketitle
\begin{abstract}
End-to-end research agents can now produce complete scientific papers, yet manuscript claims often diverge from executed experiments. This gap is structural: research state, failure histories, and claim--evidence alignment are not maintained as persistent, verifiable state across long-horizon pipelines. We present \textbf{YouRA (Your Research Agent)}, an architecture for stateful, evidence-traceable autonomous research. YouRA preserves research state, execution evidence, and failure history across the research trajectory by integrating three components: a \textit{Verification State Architecture} (VSA) that tracks hypotheses, gates, and evidence pointers; an \textit{Independent Controller} that turns state and reflection records into lifecycle, recovery, and debate/review control while separating control from execution; and \textit{Stateful Reflection} that logs failures as structured lessons and routes recovery through bounded repair, redesign, or reset. On MLR-Bench's predefined ten-task end-to-end subset, YouRA improves over both MLR-Agent and AI~Scientist~V2 on scalar Overall across all three matched backbones. An automated diagnostic using MLR-Bench's hallucination taxonomy reports intersection/union counts for four fact-based failure types, and data-provenance diagnostic shows more real-data-based outputs. Ablating each of the four components (the VSA, the Independent Controller, MCP tool access, and reflection-guided recovery) supports their separable contributions. Removing either core-state component drops YouRA below the full system. Code: \url{https://github.com/PrayPrey/Your-Research-Agent}.
\end{abstract}

\section{Introduction}\label{sec:intro}

Autonomous research agents that formulate hypotheses, run experiments, and write papers have been pursued for decades, from closed-loop experimentation \citep{king2009automation} through proposals for AI-amplified discovery \citep{gil2014amplify} to explicit Nobel-level challenges \citep{kitano2021nobel}. Large language models have made this vision practical: AI~Scientist \citep{lu2024aiscientist} and its tree-search successor V2 \citep{yamada2025aiscientistv2} automate the full pipeline from ideation to paper writing, AI-Researcher \citep{tang2025airesearcher} demonstrates role-separated multi-agent collaboration, Agent Laboratory \citep{schmidgall2025agentlaboratory} integrates human feedback, and VIRSCI \citep{su2025virsci} and AI Co-scientist \citep{gottweis2025aicoscientist} refine ideas through multi-agent debate. Context-management strategies differ across these systems, but none externalizes the research trajectory as explicit, persistent, verifiable state spanning the full pipeline. That gap limits reliability over long horizons.

This gap appears in three concrete limitations on the reliability of automatically generated research. (1)~\textbf{Context and evidence fragmentation.} When research state is implicitly embedded in conversation histories, it is vulnerable to truncation and hallucination, which breaks long-horizon coherence across hypotheses, experiments, conclusions, and manuscript claims \citep{yan2026faithfulness}. Current agents also provide only limited mechanisms for ensuring that manuscript claims remain aligned with executed code, result files, and logs. As a result, a paper can contain superficially persuasive claims generated from hallucinated results or failed runs, and such claims may still appear strong under output-only judging that does not inspect the underlying artifacts. (2)~\textbf{Lack of structured failure memory.} Agents lack mechanisms to record, diagnose, and learn from experimental failures across stages. MLR-Bench's hallucination analysis reports that evaluated coding agents produced fabricated or unverified experimental results in roughly 80\% of cases, with case studies tracing such failures to execution errors masked by simulated outputs \citep{chen2025mlrbench}. (3)~\textbf{Unreliable workflow control.} In long-horizon research agents, the same task-execution dialogue often both performs the work and decides how the workflow should continue. It carries code-writing and paper-writing context together with control decisions such as when to stop, retry, redesign a hypothesis, or move to manuscript review. As this dialogue grows, these control decisions become vulnerable to truncation, drift, and inconsistent recall, making it difficult to resume or recover a run from the actual research state.

All three limitations trace back to the same absence: the research trajectory is not maintained as explicit, persistent, verifiable structure across the pipeline. Two pieces are missing. One is a shared state substrate that records hypotheses, gates, failures, and evidence pointers. The other is an external control surface that uses this state to resume, redirect, and review the research lifecycle.

\begin{figure*}[t]
    \centering
    \includegraphics[width=0.95\textwidth]{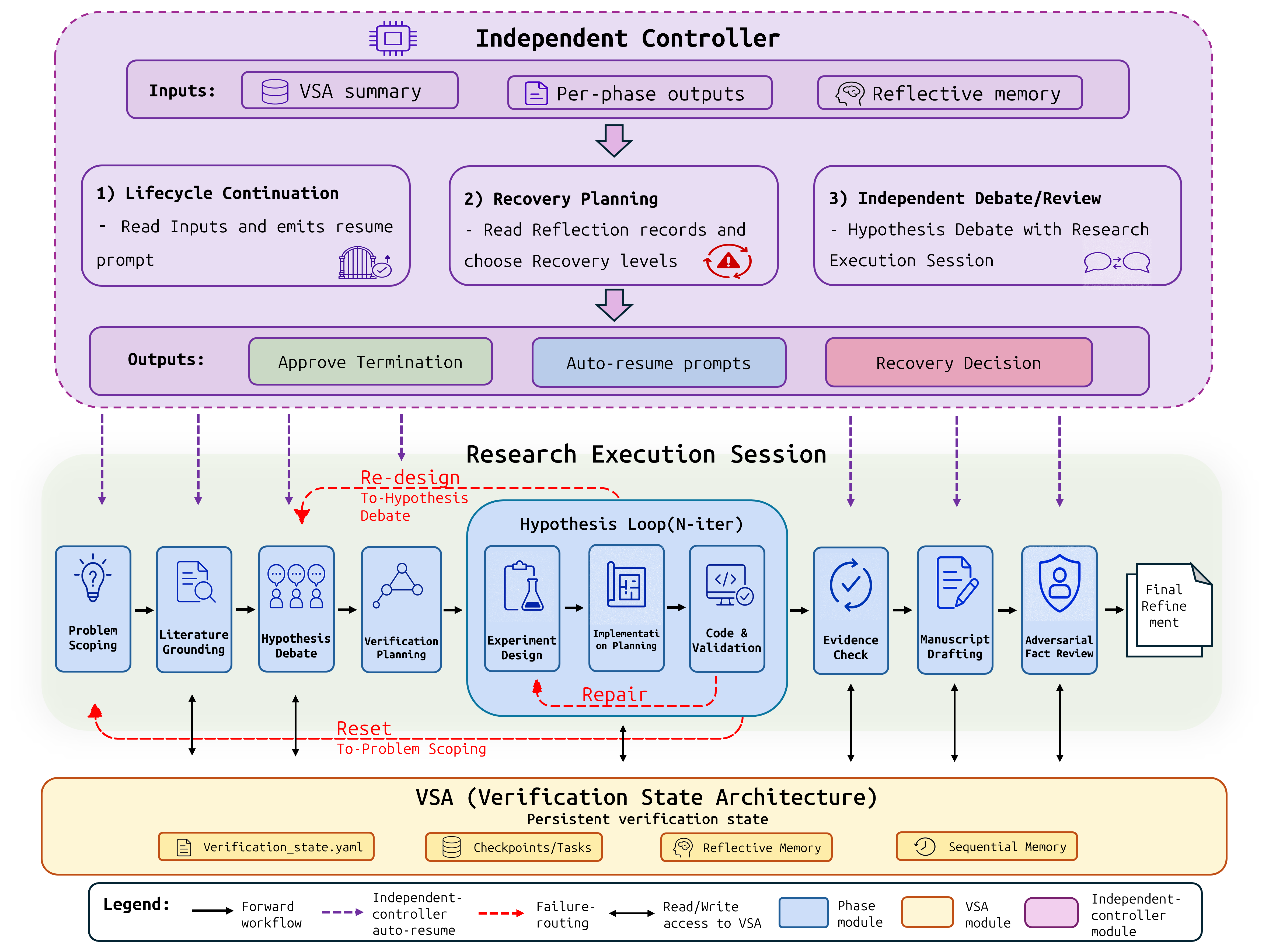}
    \caption{YouRA research lifecycle. Stages proceed left to right under the \textit{Verification State Architecture (VSA)} (\S\ref{sec:state}). An independent controller LLM manages lifecycle transitions and persona selection. Sub-hypotheses traverse the hypothesis loop, and dashed arrows denote reflection-guided three-level routing (repair, redesign, and reset) for failed gates (\S\ref{sec:collaboration}). The figure uses the reader-facing stage names introduced in \S\ref{sec:lifecycle}.}
    \label{fig:overview}
\end{figure*}

To address these problems, we present \textbf{YouRA (Your Research Agent)}, an architecture for stateful, evidence-traceable autonomous research that preserves research state, execution evidence, and trajectory-level failure history as explicit, persistent structure (Figure~\ref{fig:overview}). Its three components are: (i)~a \textbf{Verification State Architecture (VSA)} that centrally manages hypotheses, gates, and evidence pointers; (ii)~an \textbf{Independent Controller}, which consumes VSA and reflection records to drive lifecycle transitions, recovery escalation, and multi-agent debate/review roles while separating high-level control from task execution; and (iii)~\textbf{Stateful Reflection}, which logs failures as structured lessons and routes recovery through bounded repair, redesign, or reset. Given only a topic-level description, YouRA autonomously expands it into a research question, hypothesis, experiment plan, implementation, validation evidence, and final manuscript. In practical research use, the same architecture supports dynamic researcher steering through stage-specific CLI commands for redirecting, restarting, or redesigning a run; for evaluation as an autonomous research agent, however, all benchmark runs in this paper are held to fully autonomous mode.

We evaluate on MLR-Bench \citep{chen2025mlrbench}, a benchmark designed for open-ended ML research agents that provides 201 workshop-derived research tasks, stepwise and end-to-end evaluation protocols, and structured MLR-Judge rubrics. It directly matches YouRA's problem setting because it evaluates the full path from ideation through experimentation to paper writing, and explicitly surfaces fabricated or unverified experimental results as a central reliability failure. Following the benchmark's execution-heavy paper-level setting, our end-to-end and pairwise comparisons use its predefined ten-task subset. On this subset, YouRA obtains the highest 4-judge end-to-end Overall among all systems on every matched backbone, outperforming both MLR-Agent and AI~Scientist~V2 \citep{yamada2025aiscientistv2}. Order-controlled pairwise preferences also favor YouRA over both baselines on every reported backbone. Beyond the rankings, an automated diagnostic based on MLR-Bench's hallucination taxonomy decomposes concrete failure modes across matched backbones, and the data-provenance diagnostic across matched backbones shows that YouRA outputs more often rely on real data. Ablations of each of the four components support their separable contributions. Ablations of all four components significantly reduce Overall, while pairwise comparisons of their drop magnitudes do not support a reliable ranking among components.

YouRA's contribution is to make research state, execution evidence, and the research trajectory explicit, durable objects in autonomous research rather than byproducts of transient dialogue context: hypotheses, failures, evidence pointers, and recovery decisions persist in shared state and Reflective Memory, where later agents can read them. VSA-based state externalization supplies the record, the Independent Controller resumes, redirects, and reviews the lifecycle from that record while keeping control separate from execution, and Stateful Reflection drives failure-based recovery. YouRA also carries an evaluation perspective: an autonomous research system should be judged on whether its final claims remain traceable to executed evidence and the research trajectory that produced them, and a persuasive manuscript alone should not settle that judgment. 

\section{Related Work}\label{sec:related}

\paragraph{End-to-End Autonomous Research.}
Recent end-to-end systems fall along three architectural axes. \textit{Single-pipeline} designs chain ideation through paper writing into a fixed flow \cite{lu2024aiscientist}; V2 extends exploration depth via agentic tree search \cite{yamada2025aiscientistv2}. \textit{Role-separated multi-agent} designs specialize agents per stage \cite{tang2025airesearcher}, but do not make a shared persistent state the coordinating substrate for long-horizon coherence. \textit{Human-feedback-integrated} designs accept researcher feedback at stage transitions \cite{schmidgall2025agentlaboratory}, but restrict intervention to predefined checkpoints. Across these approaches, the research trajectory is still not carried between stages as durable, verifiable state, and failures never become structured constraints on later work. This is the architectural gap behind the three limitations identified in the Introduction.

\paragraph{Collaborative Systems and Persistent State.}
A parallel line of work explores human--AI collaboration for research quality. \textit{VIRSCI} \cite{su2025virsci} refines hypotheses through team-based multi-agent discussion, and \textit{AI Co-scientist} \cite{gottweis2025aicoscientist} combines Elo-ranked tournament evolution with researcher guidance, yet both stop at the pre-experimentation ideation stage. Enterprise Deep Research \cite{prabhakar2025edr} introduces steerable multi-agent deep research with human-in-the-loop guidance for analytical report generation, but targets information synthesis and does not run hypothesis-testing pipelines with code execution. YouRA targets a different setting: hypothesis-testing pipelines where state must be preserved across ideation, implementation, failure recovery, and writing. The most directly relevant mechanism for learning from failures is Reflexion \cite{shinn2023reflexion}, which applies per-task verbal reinforcement within an episodic memory buffer scoped to a single task. YouRA extends this principle to a structurally different regime. Failure lessons persist across heterogeneous research stages, from hypothesis redesign to full research-question reset, and are coordinated through the VSA instead of being held in a single-task buffer. For hypothesis generation, YouRA builds on the multi-agent scientific-debate paradigm from VIRSCI~\cite{su2025virsci} and AI Co-scientist~\cite{gottweis2025aicoscientist}, but couples it to VSA state and Reflective Memory so debate outputs remain tied to downstream verification.

\paragraph{Evaluation and Reliability.}
Research-agent reliability benchmarks now span multiple scopes. MLAgentBench \cite{huang2024mlagentbench} measures ML experimentation success. DeepResearch Bench \cite{du2025deepresearchbench} targets report quality; FML-bench \cite{zou2026fmlbenchbenchmarkingmachinelearning} evaluates exploration breadth; and MLR-Bench \cite{chen2025mlrbench} assesses the full ML-research pipeline from ideation to paper writing. MLR-Bench demonstrates that agents can mask execution failures with simulated or placeholder results, exposing a gap between fluent paper generation and trustworthy experimentation. Failure analyses trace related reliability problems to weak factual grounding and insufficient intermediate verification \cite{zhang2025genuinely}. YouRA targets this gap: it preserves failures as structured records that constrain subsequent stages, and embeds evidence-checking mechanisms before and after paper generation (\S\ref{sec:method}).

\paragraph{Limitations of Output-Only Benchmarks.}
Output-only evaluation of final papers has known limits: reference-free LLM judges show high agreement with human experts only on questions the judges themselves can answer \citep{krumdick2025nofreelabels}, paper--code alignment benchmarks reveal that even top models detect fewer than half of real discrepancies between papers and their implementations \citep{baumgartner2026scicoqa}, and coding agents lose implementation fidelity when task specifications emerge incrementally over long horizons \citep{yan2026faithfulness}. These limitations connect directly to the first problem raised in the Introduction, context and evidence fragmentation: protocols that evaluate only the final paper can structurally miss gaps between manuscript claims and executed experiments \citep{liu2026ara}. We therefore use MLR-Bench's hallucination taxonomy to inspect concrete fact-based failure types in final papers, and add a separate data-provenance diagnostic that summarizes whether manuscripts rely on real, synthetic, or fabricated artifacts. These diagnostics complement the primary benchmark rankings by exposing claim--evidence audit signals that scalar output scores can miss (\S\ref{sec:discussion}).

\section{Methodology}\label{sec:method}

Having identified the absence of persistent research state as the common root of these limitations (\S\ref{sec:intro}--\ref{sec:related}), we now describe how YouRA addresses each one through a dedicated architectural component. These structural mechanisms are intrinsic to YouRA's architecture and cannot be grafted onto existing research agents as standalone modules; the individual effect of each component is analyzed in detail in the ablation studies (\S\ref{sec:analysis}).
Here, \textit{state externalization} means storing research state (hypotheses, gates, failures, and evidence pointers) in structured files instead of conversation history, so each lifecycle stage starts from a stable record. YouRA implements this through three components that together address the three limitations from \S\ref{sec:intro}: the \textit{Verification State Architecture} (\S\ref{sec:state}) addresses context and evidence fragmentation by preserving hypotheses, gates, and evidence pointers; the \textit{Independent Controller} addresses unreliable workflow control by consuming VSA and reflection records to drive lifecycle transitions, recovery routing, and multi-agent hypothesis debate (\S\ref{sec:orchestration}); and \textit{Stateful Reflection} (\S\ref{sec:collaboration}) addresses the lack of structured failure memory through three-level failure routing. In implementation, procedural context such as reusable stage templates, reference documents, and modular skills is linked by explicit file pointers, so each stage loads the relevant context on demand. Figure~\ref{fig:overview} illustrates the full lifecycle, and \S\ref{sec:lifecycle} summarizes the stage sequence, including the evidence-checking and refinement passes that connect experimental execution to manuscript writing.

\subsection{Collaborative Research Lifecycle Overview}\label{sec:lifecycle}
The workflow comprises ten named stages under VSA-controlled transitions (Figure~\ref{fig:overview}). Problem Scoping and Literature Grounding turn a broad interest into grounded research gaps; Hypothesis Debate uses a six-persona debate protocol to converge on a hypothesis; and Verification Planning converts that hypothesis into a dependency-aware directed acyclic graph (DAG) plan and gated sub-hypotheses. The subsequent \textit{Hypothesis Loop} iterates through Experiment Design, Implementation Planning, and Coding \& Validation, turning each sub-hypothesis into an experiment specification, implementation plan, and validated code. Evidence Check records which predictions and assumptions are consistent with executed results before Manuscript Drafting uses a section-grouped drafting protocol (Story Group Architecture; Appendix~\ref{sec:appendix}) to produce the paper. Adversarial Fact Review checks the generated manuscript against the VSA and accumulated stage documents, and a final Manuscript Refinement pass integrates the review output with prior artifacts, results, logs, and code to improve logical consistency, evidential alignment, and narrative clarity. Stage-level mechanics and recovery decision logic are detailed in Appendix~\ref{sec:appendix}.

\subsection{Verification State Architecture}\label{sec:state}
YouRA's first component targets context and evidence fragmentation. In conventional agent loops, state is implicitly embedded in conversation history, where it is vulnerable to truncation and hallucination. YouRA instead maintains research state through a YAML-based \textit{Verification State Architecture (VSA)}. A central \texttt{verification\_state.yaml} tracks macro pipeline state (hypotheses, gate results, routing decisions, evidence pointers, and a failed-hypotheses registry), and per-stage checkpoint files complement it. Later stages read gate outcomes, evidence pointers, and failure records from the VSA, so executed experiments become explicit constraints when hypotheses are revised, validated, or written into manuscript claims. The VSA is a \textit{single persistent state record} under three invariants: \textit{gate-driven progression} (a hypothesis advances only when its \texttt{MUST\_WORK} gates are satisfied), \textit{auditable history} (every mutation is timestamped and attributed), and \textit{steering as mutation} (researcher directives become VSA field updates).

\subsection{Independent Controller}\label{sec:orchestration}
YouRA's second component turns persistent state into control. The VSA stores the research trajectory, and Stateful Reflection records failure constraints and routing history; the independent controller consumes these records to choose the next action. Concretely, the controller supports three roles: \textbf{lifecycle continuation}, selecting the next stage from VSA gate status; \textbf{recovery escalation}, consuming Stateful Reflection records to choose repair, redesign, or reset boundaries; and \textbf{independent debate/review}, used by Hypothesis Debate and Adversarial Fact Review. By keeping control in a separate context from task execution, YouRA avoids making the execution conversation itself the only source of control state. The controller therefore reads compact VSA/artifact summaries rather than the full trace and emits continuation context at stop or human-in-the-loop boundaries.

Prior systems such as VIRSCI and AI Co-scientist provide functional precedents for multi-agent scientific debate in hypothesis generation~\citep{su2025virsci,gottweis2025aicoscientist}. YouRA does not adopt their pipelines; it implements the analogous debate/review function inside its own stateful architecture, through a Hypothesis Debate protocol tied to the VSA and Reflective Memory. Six research personas debate candidate hypotheses while the controller LLM and task-execution LLM alternate persona contributions; Reflective Memory turns refuted approaches into constraints and validated or partially validated elements into reusable evidence. The debate terminates once specificity, mechanism clarity, testable predictions, novelty, feasibility, and objection resolution are jointly satisfied; the converged hypothesis is formalized and decomposed into 3--7 gated sub-hypotheses in the VSA.

\subsection{Stateful Reflection}\label{sec:collaboration}
YouRA's third component, \textbf{Stateful Reflection}, turns failure signals into structured constraints that later stages can use. It operates over Reflective Memory (\S\ref{sec:state}) during the hypothesis loop: routing-relevant outcomes are recorded with structured reasons, supporting or refuting evidence, and pivot-or-reuse rationales, so refuted approaches become negative constraints while validated and partially validated components become reusable evidence for successor stages. Recovery proceeds in three levels. Bounded repair first handles local Coding \& Validation failures, and after each execution an independent mock-data and hard-coded-result detector checks whether reported outputs are derived from actual data processing; detected fabrications trigger repair before evidence is passed downstream. If a \texttt{MUST\_WORK} gate remains unsatisfied after repair, Stateful Reflection escalates to Hypothesis Debate redesign when mechanism revision remains recoverable, or to Problem Scoping reset when the current hypothesis contradicts executed evidence. Each sub-hypothesis terminates when its gates resolve through validation, refutation, or escalation, after which the final state is recorded and the pipeline proceeds to Evidence Check. Routing telemetry is reported in Appendix Table~\ref{tab:reflection_envelope_main} and Appendix~\ref{sec:telemetry}.

\section{Experiments}\label{sec:experiments}

\begin{table*}[!t]
\centering
\small
\setlength{\tabcolsep}{6pt}
\renewcommand{\arraystretch}{1.25}
\begin{tabular}{ll ccccc}
\toprule
\rowcolor{blue!8}
\textbf{System} & \textbf{Backbone LLM} & \textbf{Clarity} & \textbf{Novelty} & \textbf{Soundness} & \textbf{Significance} &
\textbf{Overall} \\
\midrule
\rowcolor{gray!5}
    \multirow{3}{*}{MLR-Agent}
    & Sonnet 4.5 & $7.42 \pm 0.24$ & $5.12 \pm 0.74$ & $2.48 \pm 0.61$ & $3.52 \pm 0.68$ & $3.10 \pm 0.64$ \\
    & Opus 4.5   & $7.30 \pm 0.35$ & $5.25 \pm 0.31$ & $3.05 \pm 0.96$ & $3.80 \pm 0.71$ & $3.62 \pm 0.80$ \\
    & Sonnet 4.6 & $\mathbf{7.78 \pm 0.22}$ & $\mathbf{5.95 \pm 0.44}$ & $4.05 \pm 1.19$ & $4.72 \pm 1.14$ & $4.42 \pm 1.11$ \\
    \midrule
  \rowcolor{gray!5}
\multirow{3}{*}{AI Scientist V2}
& Sonnet 4.5 & $6.72 \pm 0.43$ & $4.85 \pm 0.60$ & $3.45 \pm 1.18$ & $3.92 \pm 0.85$ & $3.62 \pm 0.84$ \\
& Opus 4.5   & $6.38 \pm 1.25$ & $5.22 \pm 0.86$ & $4.18 \pm 1.31$ & $\mathbf{4.47 \pm 1.03}$ & $4.28 \pm 1.16$ \\
& Sonnet 4.6 & $6.45 \pm 0.65$ & $5.75 \pm 0.47$ & $3.32 \pm 1.01$ & $4.22 \pm 0.95$ & $3.88 \pm 0.88$ \\
\midrule
\rowcolor{gray!5}
  \multirow{3}{*}{\textbf{YouRA}}
& Sonnet 4.5 & $\mathbf{7.50 \pm 0.49}$ & $\mathbf{5.23 \pm 0.92}$ & $\mathbf{3.98 \pm 1.74}$ & $\mathbf{4.30 \pm 1.28}$ & $\mathbf{4.20 \pm 1.45}$ \\
& Opus 4.5   & $\mathbf{7.73 \pm 0.48}$ & $\mathbf{5.33 \pm 1.03}$ & $\mathbf{4.58 \pm 1.86}$ & $4.33 \pm 1.42$ & $\mathbf{4.45 \pm 1.51}$ \\
  & Sonnet 4.6 & $7.72 \pm 0.32$ & $5.70 \pm 0.57$ & $\mathbf{4.92 \pm 1.09}$ & $\mathbf{5.05 \pm 1.00}$ & $\mathbf{5.05 \pm 0.92}$ \\

\bottomrule
\end{tabular}
\caption{MLR-Bench end-to-end scores (10 tasks; 1--10, higher is better). Each cell reports the mean $\pm$ task SD of scores averaged over four judges; bold marks the best score per matched-backbone block.}
\label{tab:e2e}
\end{table*}

\begin{table}[t]
  \centering
  \small
  \renewcommand{\arraystretch}{1.25}
  \resizebox{\columnwidth}{!}{%
  \begin{tabular}{@{}ll ccc@{}}
  \toprule
  \rowcolor{blue!8}
  \textbf{Comparison} & \textbf{Backbone} & \textbf{Win} & \textbf{Tie} & \textbf{Lose} \\
  \midrule
  \multirow{3}{*}{YouRA vs MLR-Agent}
    & Sonnet 4.5 & 25 & 14 & 1 \\
    & Opus 4.5 & 26 & 5 & 9 \\
    & Sonnet 4.6 & 21 & 14 & 5 \\
  \midrule
  \multirow{3}{*}{YouRA vs AI Scientist~V2}
    & Sonnet 4.5 & 18 & 14 & 8 \\
    & Opus 4.5 & 17 & 12 & 11 \\
    & Sonnet 4.6 & 20 & 12 & 8 \\
  \bottomrule
  \end{tabular}%
  }
  \caption{Order-controlled pairwise preferences from the YouRA perspective. Each row summarizes 40 order-collapsed judge-task verdicts (10 tasks $\times$ 4 judges); each verdict is scored in both orders, with reversals counted as ties.}
  \label{tab:pairwise}
\end{table}

\subsection{Setup}
\paragraph{Benchmark.} We evaluate on MLR-Bench \cite{chen2025mlrbench}, a benchmark for open-ended ML research agents that provides 201 workshop-derived research tasks, MLR-Judge rubrics, and the MLR-Agent baseline pipeline for idea generation, proposal formulation, experimentation, and paper writing. The 201 tasks are sourced from recent ICLR/ICML/NeurIPS workshops and are categorized by MLR-Bench into nine core ML topics: LLMs and Vision-Language Models, AI for Science, ML Theory, Trustworthy AI, Computer Vision, ML Systems, Multimodality, Reinforcement Learning, and other emerging topics. End-to-end and pairwise paper-level evaluation use the predefined ten-task subset specified by MLR-Bench (Table~\ref{tab:e2e} and Table~\ref{tab:pairwise}). As supplementary upstream context on the full task pool, Appendix Table~\ref{tab:stagewise} compares YouRA and MLR-Agent on Idea/Proposal-stage quality across all 201 tasks.

\paragraph{Systems.} We compare three systems under Claude Sonnet~4.5~\cite{anthropic2025sonnet45}, Opus~4.5~\cite{anthropic2025opus45}, and Sonnet~4.6~\cite{anthropic2026sonnet46} backbones: (1)~\textbf{MLR-Agent} \cite{chen2025mlrbench}, the reference scaffold released with MLR-Bench, (2)~\textbf{AI Scientist~V2} \cite{yamada2025aiscientistv2}, the benchmark's primary external end-to-end comparison system and (3)~\textbf{YouRA} (ours).

Following the MLR-Bench convention~\citep{chen2025mlrbench}, we run each system in its released configuration: MLR-Agent's default scaffold, which invokes AI code editors such as Codex\footnote{\url{https://github.com/openai/codex}} and Claude Code\footnote{\url{https://github.com/anthropics/claude-code}} but does not integrate a separate MCP stack; AI~Scientist~V2 with Semantic Scholar API access and YouRA with the Model Context Protocol (MCP) tool-access stack~\citep{modelcontextprotocol2025spec} summarized in Appendix Table~\ref{tab:tools}. This is a released-configuration comparison with matched research-execution backbones, not a fully tool-normalized comparison; architecture-level interpretation therefore relies on the within-YouRA ablations and the provenance/hallucination diagnostics in addition to the cross-system scores. For YouRA, the Claude model is the task-execution backbone and the independent controller is fixed as GPT-5.2~\cite{openai2025gpt52} (removed only in the w/o Independent Controller ablation. \S\ref{sec:analysis}), so Table~\ref{tab:e2e} varies execution rather than controller backbone.

\paragraph{Evaluation.} 
We adopt MLR-Bench's End-to-End rubric~\citep{chen2025mlrbench} (Clarity, Novelty, Soundness, Significance, Overall; 1--10): \textbf{Clarity} scores readability and structure, \textbf{Novelty} scores originality and advance over prior work, \textbf{Soundness} scores method appropriateness, data support, code consistency, and whether reported results are real, not fabricated, \textbf{Significance} scores whether reliable results support field-level impact, and \textbf{Overall} gives a holistic publishability-anchored judgment. We retain these evaluation criteria and adapt MLR-Judge's input handling to account for each judge LLM's context length limit. Each judge receives the task description, paper, code, result files, and experimental logs. Across all evaluated systems, when a request exceeded this limit, we progressively truncated the combined code, result files, and logs from the end and retried the request, while keeping the task description and paper in full. The MLR-Bench study reports alignment between MLR-Judge scores and human judgments, so we use this rubric as the primary standardized metric while acknowledging the general limitations of automated judging. We therefore complement scalar scores with order-controlled pairwise preferences, hallucination diagnostics, data-provenance diagnostics, and component ablations. End-to-end scoring on the ten-task subset averages four frontier-class judges from distinct model families (Gemini~3.1~Pro~\citep{gdm2026geminipro}, GPT~5.4~\citep{openai2026gpt54}, Grok~4.3~\citep{xai2026grok43}, and Opus~4.6~\citep{anthropic2026opus46}), selected so that judge capability meets or exceeds that of the generators under evaluation while limiting judge--generator overlap. Pairwise win rates use the same four judges with order-bias control: for each paper pair, we evaluate both presentation orders and count the comparison as a tie if the preferred paper changes when the order is reversed. All evaluations run fully autonomously.


\subsection{Main Results}\label{sec:main_results}

\paragraph{Upstream stages.} We first check that YouRA does not win simply by starting from better ideas. Scoring both systems' Idea and Proposal outputs on the full 201-task pool with a single Gemini~3.0~Pro judge shared across systems (Appendix~\ref{sec:stagewise_context}), we find YouRA lower or statistically indistinguishable on Overall on both backbones measured: MLR-Agent scores higher with Sonnet~4.5, significantly so at the Idea stage, and the two do not differ with Opus~4.5. Where we can compare, the end-to-end advantage reported below therefore cannot be traced to a head start at ideation or proposal writing. It emerges over the later span of the workflow, consistent with the gains arising in execution, verification, and writing rather than in idea quality.

\begin{table*}[t]
\centering
\caption{Component ablations on MLR-Bench (10-task subset, four-LLM-judge panel, mean $\pm$ sample standard deviation). The VSA, IC (Independent Controller), MCP, and Refl.\ columns indicate whether each component is enabled in a configuration. All four ablations are run on Sonnet~4.5, Sonnet~4.6, and Opus~4.5. Rows report raw mean scores. Drops from the full configuration are discussed in the text.}
\label{tab:ablation}
\footnotesize
\setlength{\tabcolsep}{5pt}
\renewcommand{\arraystretch}{1.3}
\resizebox{1.0\textwidth}{!}{%
\begin{tabular}{@{}llcccc ccccc@{}}
\toprule
\rowcolor{blue!8}
\textbf{Configuration} & \textbf{Backbone} & \textbf{VSA} & \textbf{IC} & \textbf{MCP} & \textbf{Refl.} & \textbf{Clarity} & \textbf{Novelty} & \textbf{Soundness} & \textbf{Significance} & \textbf{Overall} \\
\midrule

\rowcolor{gray!5}
\multirow{3}{*}{YouRA (full)}
& Sonnet 4.5 & \cmark & \cmark & \cmark & \cmark & $7.50 \pm 0.49$ & $5.23 \pm 0.92$ & $3.98 \pm 1.74$ & $4.30 \pm 1.28$ & $4.20 \pm 1.45$ \\
& Sonnet 4.6 & \cmark & \cmark & \cmark & \cmark & $7.72 \pm 0.32$ & $5.70 \pm 0.57$ & $4.92 \pm 1.09$ & $5.05 \pm 1.00$ & $5.05 \pm 0.92$ \\
& Opus 4.5   & \cmark & \cmark & \cmark & \cmark & $7.73 \pm 0.48$ & $5.33 \pm 1.03$ & $4.58 \pm 1.86$ & $4.33 \pm 1.42$ & $4.45 \pm 1.51$ \\

\midrule
\rowcolor{gray!5}
\multirow{3}{*}{\quad w/o VSA}
& Sonnet 4.5 & \xmark & \cmark & \cmark & \cmark & $7.17 \pm 0.46$ & $4.12 \pm 1.09$ & $2.73 \pm 1.29$ & $3.02 \pm 0.92$ & $2.88 \pm 0.94$ \\
& Sonnet 4.6 & \xmark & \cmark & \cmark & \cmark & $7.55 \pm 0.39$ & $5.40 \pm 0.72$ & $4.20 \pm 1.62$ & $4.35 \pm 1.33$ & $4.10 \pm 1.28$ \\
& Opus 4.5   & \xmark & \cmark & \cmark & \cmark & $7.47 \pm 0.34$ & $5.42 \pm 0.43$ & $3.10 \pm 1.58$ & $3.48 \pm 0.91$ & $3.33 \pm 1.14$ \\

\midrule
\rowcolor{gray!5}
\multirow{3}{*}{\quad \makecell[tl]{w/o Independent\\\hphantom{w/o }Controller}}
& Sonnet 4.5 & \cmark & \xmark & \cmark & \cmark & $7.22 \pm 0.18$ & $4.80 \pm 1.13$ & $2.25 \pm 1.01$ & $2.88 \pm 0.44$ & $2.38 \pm 0.44$ \\
& Sonnet 4.6 & \cmark & \xmark & \cmark & \cmark & $7.53 \pm 0.25$ & $5.53 \pm 0.82$ & $4.83 \pm 1.72$ & $4.83 \pm 1.39$ & $4.70 \pm 1.22$ \\
& Opus 4.5   & \cmark & \xmark & \cmark & \cmark & $7.28 \pm 0.25$ & $5.30 \pm 0.75$ & $2.92 \pm 1.28$ & $3.70 \pm 0.78$ & $3.25 \pm 1.08$ \\

\midrule
\rowcolor{gray!5}
\multirow{3}{*}{\quad w/o MCP}
& Sonnet 4.5 & \cmark & \cmark & \xmark & \cmark & $7.45 \pm 0.42$ & $4.55 \pm 0.90$ & $2.40 \pm 0.64$ & $3.15 \pm 0.39$ & $2.83 \pm 0.54$ \\
& Sonnet 4.6 & \cmark & \cmark & \xmark & \cmark & $7.53 \pm 0.30$ & $5.75 \pm 0.60$ & $4.12 \pm 1.21$ & $4.67 \pm 0.95$ & $4.42 \pm 0.80$ \\
& Opus 4.5   & \cmark & \cmark & \xmark & \cmark & $7.08 \pm 0.44$ & $4.85 \pm 0.64$ & $2.73 \pm 0.96$ & $3.33 \pm 0.54$ & $3.05 \pm 0.73$ \\

\midrule
\rowcolor{gray!5}
\multirow{3}{*}{\quad w/o Reflection}
& Sonnet 4.5 & \cmark & \cmark & \cmark & \xmark & $7.58 \pm 0.41$ & $4.25 \pm 1.05$ & $3.15 \pm 0.95$ & $3.12 \pm 0.58$ & $3.08 \pm 0.68$ \\
& Sonnet 4.6 & \cmark & \cmark & \cmark & \xmark & $7.60 \pm 0.29$ & $5.88 \pm 0.73$ & $4.28 \pm 1.26$ & $4.80 \pm 0.77$ & $4.42 \pm 0.94$ \\
& Opus 4.5   & \cmark & \cmark & \cmark & \xmark & $7.45 \pm 0.61$ & $4.97 \pm 0.75$ & $3.58 \pm 1.51$ & $3.70 \pm 1.06$ & $3.60 \pm 1.06$ \\

\bottomrule
\end{tabular}%
}
\end{table*}

\paragraph{End-to-end scores.} Table~\ref{tab:e2e} gives the scalar results of the end-to-end evaluation. YouRA is the only system that ranks first in \emph{Overall} on all three matched backbones, and the gains are concentrated in dimensions tied to research validity. Against MLR-Agent, YouRA improves Overall on every backbone and shows the largest separation in \emph{Soundness} and \emph{Significance}. It also leads AI~Scientist~V2, the strongest prior end-to-end baseline included by MLR-Bench, on Overall across backbones and achieves higher \emph{Soundness} in all three settings. Both Overall gaps are significant under a paired sign-flip permutation test over the 30 matched (backbone, task) cells (Appendix~\ref{sec:significance_test}).

\paragraph{Pairwise preferences.} Table~\ref{tab:pairwise} reports pairwise preferences, with each pair judged in both presentation orders to control position bias~\citep{zheng2023judging}. Across the three backbones (120 comparisons per baseline), YouRA wins 72, ties 33, and loses 15 against MLR-Agent, and wins 55, ties 38, and loses 27 against AI~Scientist~V2. It comes out ahead on every backbone, so the advantage holds even when judges compare two papers from the same task side by side. An automated hallucination diagnostic with human validation complements these score-based results (\S\ref{sec:human}).

\subsection{Ablation Studies}\label{sec:analysis}

Table~\ref{tab:ablation} reports four component ablations of YouRA across all three backbones, each dropping one component from the full system. \emph{w/o MCP} removes the MCP tool layer along with tool-grounded execution and structured access to memory and external artifacts. \emph{w/o Reflection} disables reflection-guided redesign and reset, so failures and partial successes no longer turn into reusable constraints and evidence. \emph{w/o VSA} swaps the shared persistent record of hypotheses, evidence, gate outcomes, and failure constraints for a short in-context state summary at each stage. \emph{w/o Independent Controller} moves stage progression, recovery, and debate convergence into the research execution session, where they become simple artifact checks. The drops are largest on \emph{Soundness} and \emph{Significance} and smaller on \emph{Clarity} and \emph{Novelty}, which means the components matter mainly for research validity, not writing clarity or assessed novelty. Removing either core-state component, the VSA or the Independent Controller, also pushes YouRA's Overall below both baselines on Sonnet~4.5 and Opus~4.5 (Tables~\ref{tab:e2e} and~\ref{tab:ablation}).

Across the 30 matched backbone--task cells, every ablation significantly lowers \emph{Overall} under Holm-corrected paired sign-flip permutation tests. (Appendix~\ref{sec:ablation}) Comparing the drop magnitudes pairwise does not turn up statistically reliable differences among the components. These results support a contribution from each component but do not rank them against one another.


\section{Discussion}\label{sec:discussion}

\begin{table*}[t]
  \centering
  \small
  \renewcommand{\arraystretch}{1.5}
  \begin{tabular}{@{}lcc@{\hspace{1.2em}}lcc@{\hspace{1.2em}}lcc@{}}
  \toprule
  \rowcolor{blue!8}
  \multicolumn{3}{@{}l}{\textbf{Flag precision by category}} & \multicolumn{3}{l}{\textbf{Flag precision by system}} & \multicolumn{3}{l@{}}{\textbf{Auditor agreement}} \\
  \rowcolor{blue!8}
  & \textbf{Count} & \textbf{Rate} & & \textbf{Count} & \textbf{Rate} & & \textbf{Count} & \textbf{Rate} \\
  \midrule
  Hallucinated Methodology & 92/102 & 90.2\% & YouRA & 68/90 & 75.6\% & YouRA & 26/30 & 86.7\% \\
  Faked Experimental Results & 64/85 & 75.3\% & MLR-Agent & 70/90 & 77.8\% & MLR-Agent & 21/30 & 70.0\% \\
  Mathematical Errors & 15/21 & 71.4\% & AI Scientist V2 & 65/90 & 72.2\% & AI Scientist V2 & 22/30 & 73.3\% \\
  Nonexistent Citations & 32/62 & 51.6\% & & & & & & \\
  \midrule
  \textbf{Overall} & \textbf{203/270} & \textbf{75.2\%} & \multicolumn{3}{l}{$\chi^2(2)=0.75$, $p=0.686$} & \textbf{Overall} & \textbf{69/90} & \textbf{76.7\%} \\
  Wilson 95\% CI & \multicolumn{2}{c}{[69.7\%, 80.0\%]} & & & & Cohen's $\kappa$ & \multicolumn{2}{c@{}}{0.54} \\
  \bottomrule
  \end{tabular}
  \caption{Human validation of the automated hallucination diagnostic. Left: precision of automatically flagged instances by category (confirmed/flagged), with the Wilson 95\% CI of the overall precision. Middle: per-system precision; the $\chi^2$ test shows the differences across systems are not significant. Right: agreement between the three primary annotators and an independent fourth auditor on a 90-instance subset (30 per annotator), with Cohen's $\kappa$; 18 of the 21 disagreements are cases where the auditor is stricter.}
  \label{tab:human_validation}
\end{table*}

\begin{table}[t]
  \centering
  \small
  \renewcommand{\arraystretch}{1.25}
  \resizebox{\columnwidth}{!}{%
  \begin{tabular}{@{}ll ccc@{}}
  \toprule
  \rowcolor{blue!8}
  & \textbf{Flags} & \textbf{Precision} & \textbf{Corrected} \\
  \midrule
  YouRA & 420 & 75.6\% & 317 \\
  MLR-Agent & 552 & 77.8\% & 429 \\
  AI Scientist V2 & 413 & 72.2\% & 298 \\
  \midrule
  \rowcolor{blue!8}
  & $\Delta$ & \textbf{95\% CI} & $p$ \\
  \midrule
  YouRA $-$ MLR-Agent & $-112$ & $[-184, -39]$ & 0.003 \\
  YouRA $-$ AI Scientist V2 & $+19$ & $[-45, +86]$ & 0.571 \\
  AI Scientist V2 $-$ MLR-Agent & $-131$ & $[-200, -62]$ & ${<}0.001$ \\
  \bottomrule
  \end{tabular}
  }
    \caption{Precision-corrected hallucination counts, computed as automated flags times the human-validated precision from Table~\ref{tab:human_validation}, with pairwise differences (95\% CIs, two-sided $p$).}
  \label{tab:corrected_counts}
\end{table}

\subsection{Human validation}\label{sec:human}
The preceding analysis shows that YouRA's mechanisms were exercised in evaluation and that component ablations support their contribution to scalar performance. We next examine how this advantage appears in the failure modes tracked by MLR-Bench: faked experimental results, hallucinated methodology, nonexistent or incorrect citations, and mathematical errors~\citep{chen2025mlrbench}. For papers generated across the three matched backbones, we first apply the MLR-Judge automated diagnostic using four judges. We report union and intersection counts per category: the former count papers (out of ten per system and backbone) in which at least one judge flagged that category, whereas the latter count only papers flagged by all four judges (Appendix Figure~\ref{fig:hallucination_taxonomy_bounds}). We then conduct a human validation of the automatically flagged instances, summarized in Table~\ref{tab:human_validation}. Three evaluators independently assess 270 flagged instances (90 per system), confirming roughly three quarters as genuine hallucinations. Precision is highest for hallucinated methodology and lowest for nonexistent citations, and per-system precision is statistically indistinguishable, so the automated diagnostic does not systematically favor or disfavor any system. A fourth evaluator then audits the three annotators by independently re-annotating a 90-instance subset. Agreement is moderate, most disagreements are cases where the auditor is stricter than the primary annotator, and the disagreement rate does not differ significantly across systems ($\chi^2(2)=2.61$, $p=0.27$), giving no evidence of systematic bias toward or against any of the three systems.

Because precision is comparable across systems, Table~\ref{tab:corrected_counts} reports precision-corrected hallucination counts. On these corrected counts, MLR-Agent is significantly worse than both other systems, while YouRA and AI~Scientist~V2 are not statistically distinguishable; this is absence of evidence for a difference, not evidence of equivalence. On Overall, in contrast, YouRA is significantly better than both baselines (\S\ref{sec:main_results}). YouRA thus sits in the low-hallucination group yet alone leads on Overall, so its score advantage is not achieved through more fabricated content.

\subsection{Data provenance}\label{sec:data_provenance}

\begin{figure}[t]
\centering
\includegraphics[width=\columnwidth]{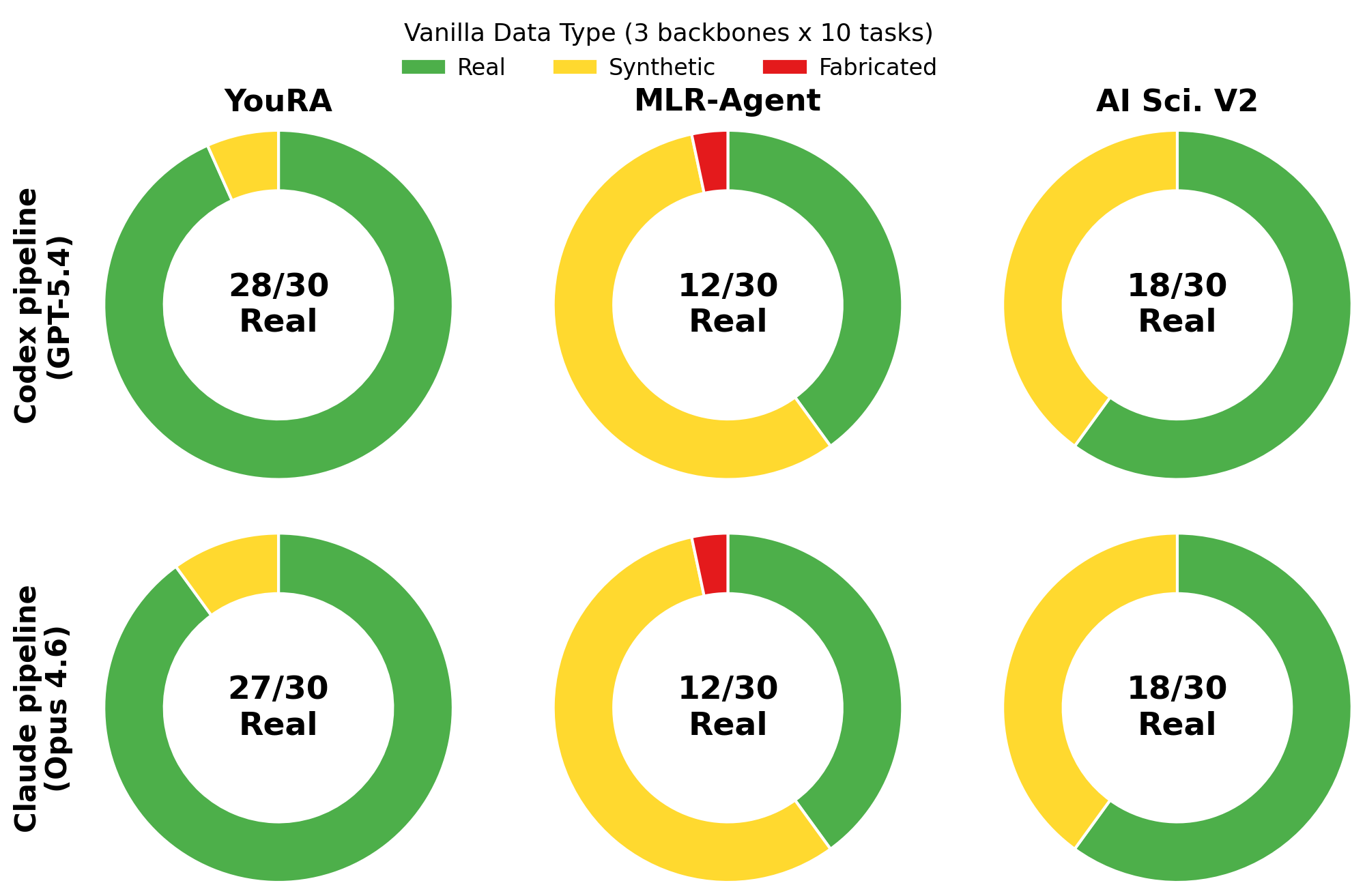}
\caption{Data-provenance diagnostics on the Sonnet~4.5, Sonnet~4.6, Opus~4.5 tasks subset. Each donut summarizes the Real/Synthetic/Fabricated classification of 30 generated papers for one system under one of two independent analysis pipelines (Codex/GPT-5.4 or Claude/Opus~4.6). Exact counts are in Appendix Table~\ref{tab:provenance_backbone}.}
\label{fig:vanilla_data_type}
\end{figure}

The data-provenance companion in Figure~\ref{fig:vanilla_data_type} and Appendix Table~\ref{tab:provenance_backbone} provides a separate provenance context. In Sonnet 4.5, Sonnet 4.6 and Opus 4.5, 27/30 of YouRA's generated papers are classified as real-data-based by both analyzers, compared with 11/30 for MLR-Agent and 17/30 for AI~Scientist~V2. These diagnostics complement the score and preference results in Table~\ref{tab:e2e} and Table~\ref{tab:pairwise} with concrete failure-type and data-provenance views.

\section{Conclusion}\label{sec:conclusion}
YouRA addresses a persistent gap in autonomous research agents, the drift of manuscript claims away from executed experiments, by making research state, execution evidence, and the research trajectory explicit and durable across the pipeline. VSA-based state externalization, the Independent Controller, and Stateful Reflection record hypotheses, failures, evidence links, and recovery decisions as persistent structures, and removing any one of them significantly lowers end-to-end quality. On MLR-Bench, YouRA improves scalar end-to-end scores over both baselines across matched backbones. Human-validated hallucination counts on the final papers, checked against code and logs, and a provenance classification of the data behind each run's experiments together show that the gain does not rest on fabricated content. YouRA has significantly fewer confirmed hallucinations than MLR-Agent, its count cannot be distinguished from that of AI~Scientist~V2, its runs rely on real data more often than either baseline under a conservative reading, and the scalar score separates YouRA from both. A scalar end-to-end score cannot by itself tell whether a gain comes from better research or from more persuasive fabrication, so it should be paired with diagnostics that check manuscript claims and their underlying data against executed evidence, such as the human-validated hallucination diagnostic and the automated data-provenance analysis used here.

\section*{Limitations}

All experiments focus on machine learning and Deep learning tasks, leaving the applicability of YouRA to other scientific domains for future investigation. Research quality is assessed using the MLR-Judge framework with a judge LLM different from those evaluated in the original MLR-Bench study. Alignment with human reviewers therefore requires separate validation for our judge configuration. Input truncation due to the judge LLMs' context length limits may disadvantage papers with extensive code, result files, or logs. The systems compared in our experiments operate with different tool configurations and token budgets, so the reported scores reflect performance under each system's resource conditions. A promising direction for future work is to develop a benchmark with standardized tool access and token budgets, providing a common experimental setting for comparing autonomous research systems under matched resource constraints.

\section*{Ethical Considerations}

Autonomous paper-generation systems can accelerate legitimate research workflows, but they also create risks of unverified submissions, misleading claims, and over-reliance on AI-generated manuscripts. YouRA is intended to assist researchers, not replace them, so all outputs should be independently verified and AI involvement in manuscript preparation should be disclosed. Because the system uses commercial LLM APIs and external tools, deployments should follow institutional and data-use policies before sending private code, data, or unpublished research ideas to external services; such dependencies may also limit accessibility for under-resourced groups, motivating future support for open-weight or local alternatives. We use MLR-Bench tasks under their stated license and document external-tool usage policies in the released repository. The hallucination and data-provenance diagnostics in this paper are audit tools for checking claim--evidence alignment, not methods for polishing unsupported manuscripts; responsible use requires preserving artifact provenance and making claim-to-evidence links inspectable.

\section*{Acknowledgements}
This work was supported by Institute of Information and Communications Technology Planning and Evaluation(IITP) grant funded by the Korea government(MSIT) (RS-2019-II190004, Development of semi-supervised learning language intelligence technology and Korean tutoring service for foreigners).

\bibliography{custom}

\appendix

\section{System Implementation Details}\label{sec:appendix}

This appendix explains how the VSA, Independent Controller, and Stateful Reflection mechanisms introduced in the main methodology are implemented in the deployed workflow. The implementation is organized into four layers: external tool integration, state externalization mechanisms for long-horizon persistence, a centralized state file shared by all agents, and stage-specific agent assignments.

\paragraph{External Tool Integration.}
YouRA uses Claude Code as the execution host and connects the five MCP-facing tools in Table~\ref{tab:tools} as a common tool-access layer. \textbf{Archon} provides retrieval-augmented generation (RAG)-backed knowledge management, storing project-specific documents and prior implementation cases in a vector database. \textbf{Exa} offers Search, Contents, and Research APIs for web-based evidence collection. \textbf{Semantic Scholar} supplies Academic Graph and Recommendations APIs for paper metadata and citation networks. \textbf{Serena} enables symbol-level code navigation and editing, with failure-context recording for Reflective Memory. \textbf{Clear Thought} provides structured reasoning scaffolds, including scientific-method and mental-model templates. Claude Code hooks provide termination control and iterative execution loops; Table~\ref{tab:tools} lists only the external tool services exposed through MCP-style access.

\begin{table*}[t]
    \centering
    \small
    \renewcommand{\arraystretch}{1.25}
    \begin{tabularx}{\textwidth}{@{}l l l X@{}}
    \toprule
    \rowcolor{blue!8}
    \textbf{Tool} & \textbf{Role} & \textbf{Lifecycle coverage} & \textbf{Usage in YouRA} \\
    \midrule
    \rowcolor{gray!5}
    Archon           & Sequential Memory & Full lifecycle & RAG-backed knowledge base; stores implementation patterns; manages task lifecycle (todo/doing/review/done) \\
    Exa              & Evidence Search   & Grounding, debate, planning, validation & Web-based evidence collection via Search, Contents, and Research APIs \\
    \rowcolor{gray!5}
    Semantic Scholar & Literature Search & Grounding, debate, evidence, writing, review & Academic Graph and Recommendations APIs for paper metadata and citations \\
    Serena           & Code Analysis / Memory & Scoping, debate, hypothesis loop, writing, review & Code-aware symbol navigation; optional in Experiment Design when base code exists, stronger in Implementation Planning--Coding \& Validation; memory support where explicitly invoked \\
    \rowcolor{gray!5}
    Clear Thought    & Structured Reasoning & Debate and evidence checking & Scientific method and mental models for hypothesis refinement \\
    \bottomrule
    \end{tabularx}
    \caption{MCP-facing tool-access stack used by YouRA. Claude Code is the execution host and hook surface and has no row of its own; the rows list external tool services exposed through MCP-style access. Tool URLs: Archon (\url{https://github.com/coleam00/Archon/tree/archive/v1-task-management-rag}), Exa (\url{https://exa.ai}), Semantic Scholar (\url{https://www.semanticscholar.org/product/api}), Serena (\url{https://github.com/oraios/serena}), Clear Thought (\url{https://github.com/Dicklesworthstone/mcp-clear-thought}).}
    \label{tab:tools}
\end{table*}

\paragraph{State Externalization Mechanisms.}
MCP tool calls are stateless, so YouRA separates transient tool calls from durable artifacts. BMAD-Method\footnote{\url{https://github.com/bmad-code-org/bmad-method}} provides structured artifact templates so each stage receives the prior stage's outputs in a fixed schema. During Problem Scoping, YouRA adapts BMAD's brainstorming workflow as a Socratic Discovery procedure for topic expansion: a topic-level prompt is expanded into diverse research angles through iterative questioning, anti-bias pivots, and convergence into a research-question package passed to Literature Grounding. Archon's RAG-backed knowledge management stores project-specific documents, code patterns, and prior implementation cases in a vector database, which agents query during Implementation Planning and Coding \& Validation. Agent Skills\footnote{\url{https://agentskills.io/specification}} loads procedural knowledge on demand, including prompt templates, verification protocols, and output schemas. These mechanisms are supporting infrastructure around the VSA: they keep state available outside the transient conversation context.

\paragraph{Verification State Architecture.}
The VSA invariants are introduced in \S\ref{sec:state}. Every agent reads from and writes to the centralized YAML file, and Figure~\ref{fig:vsa} shows a simplified schema. The schema excerpt preserves implementation labels used for machine routing; the surrounding prose uses reader-facing stage names.

\begin{figure}[!t]
\begin{lstlisting}[basicstyle=\ttfamily\tiny,columns=fullflexible]
metadata:
project_name: "..."
main_hypothesis_id: H-main
schema_version: "3.5"

workflow:
status: ACTIVE
current_phase: Phase 4.5
next_action: "..."
failure_routing:
  phase4_must_work_partial:
    route_after_max: Phase 2A-Dialogue
  phase4_must_work_fail:
    route_to: Phase 0

sub_hypotheses:
h-e1:
  status: COMPLETED
  gate: {type: MUST_WORK, result: PASS}
  experiment_design:
    file: h-e1/02c_experiment_brief.md
  implementation_planning:
    tasks_file: h-e1/03_tasks.yaml
  validation:
    result: PASS
    key_findings: ["..."]

history:
- event: Phase 4 PASS - h-e1 validated
timestamp: "..."

statistics:
validated_sub_hypotheses: 3
gates_passed: 3

\end{lstlisting}
\caption{Simplified excerpt of the VSA state file. \texttt{workflow} governs stage transitions and recovery routing; each
\texttt{sub\_hypotheses} entry stores gate status, artifact pointers, and validation findings; \texttt{history} records
auditable state transitions. Implementation labels are shown as stored in the machine-readable state, while the prose uses reader-facing stage names.}
\label{fig:vsa}
\end{figure}

\paragraph{Stage-Agent Assignment.}
Table~\ref{tab:agents} summarizes which agents and tools the controller invokes from each VSA stage state. Agents fall into three functional groups. \textit{Planning agents} span Problem Scoping through Experiment Design, turning interests into grounded questions, debated hypotheses, verification plans, and experiment specifications. \textit{Execution agents} cover Implementation Planning and Coding \& Validation, producing PRDs, architectures, task lists, implementations, and validation outcomes. \textit{Synthesis agents} cover Evidence Check, Manuscript Drafting, Adversarial Fact Review, and Manuscript Refinement, carrying executed evidence into manuscript claims and review. The Tools column summarizes external tool access; all stages read from and write to the VSA.

\begin{table*}[t]
    \centering
    \small
    \renewcommand{\arraystretch}{1.25}
    \begin{tabularx}{\textwidth}{@{}>{\raggedright}p{3.7cm} >{\raggedright}p{3.6cm} >{\raggedright}X >{\raggedright\arraybackslash}X@{}}
    \toprule
    \rowcolor{blue!8}
    \textbf{Workflow stage} & \textbf{Tools} & \textbf{Input} & \textbf{Output} \\
    \midrule
    \rowcolor{gray!5}
    Problem Scoping & Archon, Serena & Researcher's interests & Concrete research question \\
    Literature Grounding & Archon, SS, Exa & Research question & 3 research gaps \\
    \rowcolor{gray!5}
    Hypothesis Debate & Archon, SS, Exa, Serena, ClearThought & 3 research gaps & Formalized hypothesis \\
    Verification Planning & Archon, Exa, ClearThought & Formalized hypothesis & 3--7 sub-hypothesis DAG \\
    \rowcolor{gray!5}
    Experiment Design & Archon, Exa, Serena & Sub-hypotheses & Experiment specification \\
    Implementation Planning & Archon, Serena & Experiment spec & PRD, architecture, logic, config, tasks \\
    \rowcolor{gray!5}
    Coding \& Validation & Archon, Exa, Serena & Tasks & Validated implementation \\
    \rowcolor{gray!5}
    Evidence Check & Archon, Serena & All Coding \& Validation artifacts & Evidence-check summary \\
    Manuscript Drafting & Archon, Serena & Evidence-check summary & Draft manuscript \\
    \rowcolor{gray!5}
    Adversarial Fact Review & Archon, Serena & Manuscript + VSA + code & Reviewed paper \\
    Manuscript Refinement & -- & All artifacts & Refined final paper \\
    \bottomrule
    \end{tabularx}
    \caption{Stage-agent assignment. \textbf{SS} = Semantic Scholar. The Tools column lists external tool access; all stages read from and write to the VSA.}
    \label{tab:agents}
\end{table*}

\paragraph{Independent Controller.}
The Independent Controller turns VSA records into stage transitions, recovery choices, and persona selection, making persistent state an active control surface rather than a passive log. It runs in an independent LLM context decoupled from the task-execution agent; in our implementation, this controller is GPT-5.2 \cite{openai2025gpt52}, accessed through OpenRouter API. At each step it reads only two inputs: a compact summary of the current VSA and the most recent per-stage artifact pointer. Because the controller never sees the full execution trace, its context does not grow with the pipeline length, and its routing decisions are less directly exposed to the task-execution agent's internal reasoning. The same Independent Controller is reused in Hypothesis Debate, where it selects personas and judges convergence; in Adversarial Fact Review, the independent critics are the reviewer personas, which run as context-isolated task agents of the backbone LLM, while the controller's Stop hook handles only round-to-round continuation. Operationally, the controller's per-stage output is consumed by a Stop hook that either approves termination, emits a resume prompt back to the task-execution LLM, or escalates a \texttt{MUST\_STOP} signal when human intervention is required.

\paragraph{Reflection-Guided Failure Handling.}
When a Coding \& Validation gate fails, recovery first separates local repair from archive-producing escalation. Local runtime, test, or fabricated-output failures are handled as bounded repair inside Coding \& Validation; such attempts may appear in validator traces but do not create separate archive snapshots. Only when the gate remains unsatisfied does the reflection step classify the failure by recoverability and evidence contradiction, then select same-hypothesis retry, Hypothesis Debate redesign, or Problem Scoping reset. Table~\ref{tab:recovery} summarizes these externally visible routing levels, and Algorithm~\ref{alg:recovery_pseudocode} gives the paper-level decision logic. Reflective Memory and the VSA \texttt{recovery\_history} record routing-relevant failure context and escalated outcomes, so downstream stages inherit constraints learned from failure. Sequential Memory (Archon) records the archive-producing routing/recovery events summarized in Table~\ref{tab:reflection_envelope_main}.

\begin{table}[t]
    \centering
    \small
    \renewcommand{\arraystretch}{1.25}
    \begin{tabularx}{\columnwidth}{@{}l l X@{}}
    \toprule
    \rowcolor{blue!8}
    \textbf{Level} & \textbf{Target} & \textbf{Criterion} \\
    \midrule
    \rowcolor{gray!5}
    Repair & Coding \& Validation (in-loop) & Local runtime, test, or fabricated-output failure repairable before archive-producing escalation \\
    Redesign & Hypothesis Debate & \texttt{MUST\_WORK} failure requiring hypothesis or mechanism revision after repair \\
    \rowcolor{gray!5}
    Reset & Problem Scoping & Evidence contradiction indicating that the current hypothesis is unsupported \\
    \bottomrule
    \end{tabularx}
    \caption{Reflection-guided failure-routing levels. Bounded repair is handled in-loop and does not create an archive snapshot; redesign and reset are archive-producing escalations.}
    \label{tab:recovery}
\end{table}

\begin{algorithm}[t]
\caption{Reflection-guided failure handling in the hypothesis loop}
\label{alg:recovery_pseudocode}
\footnotesize
\begin{algorithmic}[1]
\Require Failed gate type $g$, failure context $c$, retry state $r$
\Ensure Next action $a$
\If{the failure is local and a bounded Coding \& Validation repair attempt remains}
    \State \Return bounded repair in Coding \& Validation
\EndIf
\If{$g = \texttt{SHOULD\_WORK}$}
    \If{a same-hypothesis improvement path remains available}
        \State \textsc{Same-hypothesis retry} decision 
        \State \Return retry from Experiment Design within the current hypothesis
    \Else
        \State record the limitation in Reflective Memory and the VSA
        \State \Return continue the pipeline with the limitation documented
    \EndIf
\EndIf
\State $\alpha \gets$ reflection over failure severity, recoverability, and evidence contradiction
\State write $\alpha$ to Reflective Memory and the VSA \texttt{recovery\_history}
\If{$\alpha = \textsc{Same-hypothesis retry}$}
    \State \Return retry from Hypothesis-Loop within the current hypothesis
\ElsIf{$\alpha = \textsc{Redesign}$}
    \State \Return reroute to Hypothesis Debate for hypothesis redesign
\Else
    \State \Return reroute to Problem Scoping for question reset
\EndIf
\end{algorithmic}
\end{algorithm}

\paragraph{Hypothesis Debate.}
Hypothesis Debate implements the function of multi-persona scientific debate inside YouRA's stateful workflow. The six personas cover novelty and ideation (Dr.~Nova), falsifiability and experiment design (Prof.~Vera), scientific contribution and impact (Dr.~Sage), technical/theoretical feasibility (Prof.~Pax), evidence-based hypothesis refinement (Dr.~Ally), and assumptions/failure modes (Prof.~Rex). The protocol has three stages. First, Gap Selection chooses the priority research gap and supporting evidence from Literature Grounding. During Debate, the independent controller LLM (GPT-5.2 \cite{openai2025gpt52}) reads recent dialogue and VSA state to decide convergence and select the next personas, while the backbone LLM produces the role-conditioned responses. If a network or API failure prevents a controller call, backbone-LLM fallback preserves execution continuity for that round. Result Structuring converts the debate outcome into structured YAML for downstream stages. When this stage is re-entered after a failed or superseded hypothesis, Reflective Memory is checked first so the system does not repeat refuted directions. The converged hypothesis is formalized with variables, causal mechanisms, a null hypothesis, and testable predictions, then decomposed during Verification Planning into a 3--7-node sub-hypothesis DAG with \texttt{MUST\_WORK}/\texttt{SHOULD\_WORK} gates recorded in the VSA.

\paragraph{Hypothesis Loop Details: Experiment Design through Coding \& Validation.}
Each sub-hypothesis traverses the hypothesis loop from Experiment Design through Coding \& Validation. Experiment Design turns the sub-hypothesis into an experiment specification covering the dataset, model architecture, and evaluation metrics. Implementation Planning expands this specification into PRD, architecture, logic, and configuration documents, then consolidates them into the VSA task artifact \texttt{03\_tasks.yaml}, a prioritized implementation task list whose entries store references to the relevant specification sections. Coding \& Validation uses these references to apply SDD (Spec-Driven Development): for each task, it extracts the governing requirements, generates and runs compliance tests before implementation, implements the task, and invokes an independent Validator for multi-stage verification. Failures are fed back with diagnostic context, unmet gates trigger reflection-guided routing, and a hook-based mock-data/hard-coded-result detector checks whether outputs come from actual data processing or from fabricated placeholders. Each completed sub-hypothesis is marked as validated, refuted, or escalated; the loop continues until all sub-hypotheses reach a terminal state, after which the pipeline proceeds to Evidence Check.

\paragraph{Evidence Check.}
Evidence Check produces an evidence-check summary that records, for each prediction, whether the executed result supports it (\textsc{supported}, \textsc{partially\allowbreak\_supported}, \textsc{refuted}, or \textsc{inconclusive}); the same pass also tracks explicit causal-mechanism claims and assumptions. This stage is an evidence-transfer check: it records which predictions and assumptions match executed results, then passes that summary to Manuscript Drafting and Adversarial Fact Review. The resulting evidence-check summary is a core input to writing and review alongside the VSA and Coding \& Validation artifacts; its downstream effect on manuscript-claim traceability is assessed through the hallucination and data-provenance diagnostics in Appendices~\ref{sec:hallucination_diagnostic} and ~\ref{sec:provenance}.

\paragraph{Manuscript Drafting: Story Group Architecture.}
Manuscript Drafting generates the manuscript in three sequential story groups, guided by a blueprint that is written \textit{before} any section content. \textit{Foundation} (Introduction + Related Work + Methodology) frames the problem and motivates the solution. \textit{Evidence} (Experiments + Results + Discussion) presents and interprets the experimental findings. \textit{Closure} (Conclusion + Abstract) is produced last, and the abstract is written last to capture the full paper context. The Evidence Check summary is one input to manuscript writing, alongside the VSA and artifacts from Problem Scoping through Coding \& Validation (e.g., Problem Scoping for introduction motivation, Experiment Design/Implementation Planning for methodology, Coding \& Validation for results). VSA-recorded values are the intended source for reported numbers, discouraging unsupported reported values.

\paragraph{Adversarial Fact Review and Manuscript Refinement.}
Adversarial Fact Review applies an adversarial fact-review protocol (\textit{Devil's Advocate}) with three reviewer personas. The \textit{Accuracy Checker} cross-references every claim against ground truth, compares reported numbers against a dedicated ground-truth registry, and checks methodology consistency and cross-section contradictions against Coding \& Validation artifacts. The \textit{Bored Reviewer} examines whether the abstract and framing would compel further reading from the perspective of a time-pressed conference reviewer. The \textit{Skeptical Expert} probes novelty claims and baseline fairness. Findings are classified into two blocking levels. \textit{Major} findings are claim--implementation contradictions or reviewer-attackable weaknesses such as unfair baselines that must be corrected with evidence before convergence; \textit{Minor} findings are polish issues such as typos, awkward phrasing, and formatting, collected as human-review notes. Manuscript Refinement revisits the VSA, Evidence Check summary, stage artifacts, result files, logs, and code until no major findings remain.

\section{Reproducibility Details}\label{sec:reproducibility}

This section summarizes the code/artifacts, failure-routing logic, and execution configuration needed to reproduce the core results.

\paragraph{Code and Artifacts.}
The complete source code, VSA schemas, controller configurations, agent prompts, evaluation scripts, generated papers, VSA state files, recovery snapshots, judge prompts and raw outputs, hallucination-diagnostic outputs, and table/figure regeneration scripts are released at \url{https://github.com/PrayPrey/Your-Research-Agent}.

\paragraph{Dataset and Artifact Licensing.}
MLR-Bench tasks \citep{chen2025mlrbench} are released by the original authors under their stated license, and we use them for non-commercial academic evaluation only. The reference papers, datasets, and starter code that each task points to are accessed under the licenses of their original sources and are not redistributed by us. Manuscripts and code generated by YouRA in the course of these benchmark runs are released for research use under the same terms as the YouRA repository and inherit any upstream licensing obligations from the underlying datasets they exercise; users who repurpose generated artifacts must comply with those upstream terms.

\paragraph{Failure Classification Logic.}
The Validator and reflection step together map each gate failure to one of the three externally visible routing levels in Table~\ref{tab:recovery}; Algorithm~\ref{alg:recovery_pseudocode} gives the compact paper-level decision procedure. Repair is bounded in-loop handling inside Coding \& Validation, whereas redesign and reset are escalations that write structured archive snapshots.

\begin{itemize}
\item \textbf{Repair} (runtime / test / fabricated-output failure): the implementation produces an exception, assertion error, test failure, or mock-data/hard-coded-output signal that is localizable to a specific code location. The Validator attempts bounded in-place fixes within Coding \& Validation before any archive-producing escalation.
\item \textbf{Redesign routing} (design / mechanism failure): a \texttt{MUST\_WORK} failure persists after bounded repair, but reflection judges that the failure remains recoverable through hypothesis or mechanism revision (e.g., architectural mismatch, invalid causal mechanism, dependency structure that requires redesign). The system routes to Hypothesis Debate with a diagnostic record specifying the failed component and attempted fixes.
\item \textbf{Reset} (fundamental failure): a \texttt{MUST\_WORK} failure reflects evidence that fundamentally contradicts the current hypothesis (e.g., predicted correlation does not hold, effect size remains absent after correct implementation). The system routes to Problem Scoping with the full failure context, archiving the current research folder under a timestamped backup before beginning a new research direction.
\end{itemize}

\paragraph{Execution Configuration.}
All experiments use the backbone LLM's default sampling parameters (temperature and top-p as set by the provider's API defaults). We evaluate Coding \& Validation reflection-guided routing by reporting observed archive-producing routing/recovery counts for each ten-task backbone batch in Table~\ref{tab:reflection_envelope_main} and Appendix Tables~\ref{tab:telemetry_routing_count}--\ref{tab:telemetry_routing_levels}. The six-persona debate in Hypothesis Debate runs for up to 20 rounds in our implementation (one persona contribution per round), terminating when all six convergence criteria (specificity, mechanism clarity, testable predictions, novelty, feasibility, and objection resolution) are jointly satisfied. Coding \& Validation SDD cycles iterate until gate criteria resolve through validation, refutation, or escalation. The MCP tool stack is listed in Table~\ref{tab:tools}; each server is installed from its upstream repository.

\paragraph{Reproducibility Checklist Disclosure.}
We disclose the reproducibility-relevant configuration for this version: (i)~\textbf{random seeds} are not used as a reproducibility guarantee for the controller/evaluation layer because it routes through multiple commercial LLM API paths (OpenRouter / Anthropic and provider-backed judge endpoints) that do not provide a uniform deterministic replay guarantee across models and backend updates; we use the provider-default API sampling parameters stated above. (ii)~\textbf{Hardware specification.} All experiments ran on a single workstation with the following configuration: 2$\times$ AMD EPYC 9354 32-core CPUs (total 64 physical cores / 128 threads), 128~GB system RAM, up to 5$\times$ NVIDIA H100 NVL GPUs (94~GB HBM3 each; NVIDIA driver 575.57.08), Ubuntu 22.04 LTS (Linux kernel 5.15), and Python 3.12. The controller/evaluation layer (Hypothesis Debate controller, judge calls, baseline LLM backbones) runs entirely against commercial APIs and consumes no local GPU. The Coding \& Validation \emph{execution} layer, in contrast, runs the LLM-generated PyTorch experiment code locally on the workstation's GPU(s); the actual per-task GPU count varies with the experiment generated by the agent (typically 1 GPU; multi-GPU is only used when the generated code explicitly requests it). All experiments are single-node and use no multi-node distributed training.


\section{Operational Telemetry}\label{sec:telemetry}

This appendix reports operational statistics for the three-level routing mechanism (\S\ref{sec:collaboration}), aggregated from the \texttt{verification\_state.yaml} final state and routing-recovery archive snapshots of the benchmark artifacts (10 MLR-Bench tasks $\times$ 3 backbones). Aggregation, classification, and end-to-end verification scripts are released with the repository; every value reported here was independently re-derived from raw artifacts as a 30/30 task verification check before submission. Table~\ref{tab:reflection_envelope_main} provides the compact backbone-wise summary, and the subsequent tables give per-task and per-level evidence.

\begin{table}[t]
\centering
\small
\renewcommand{\arraystretch}{1.2}
\resizebox{\columnwidth}{!}{%
\begin{tabular}{@{}l ccccc@{}}
\toprule
\rowcolor{blue!8}
\textbf{Backbone} & \textbf{Tasks} & \textbf{Total} & \textbf{Per task} & \textbf{Median} & \textbf{Range} \\
\midrule
\rowcolor{gray!5} Sonnet~4.5 & 10 & 28 & 2.8 & 2.5 & 0--8 \\
Opus~4.5            & 10 & 40 & 4.0 & 3.0 & 0--11 \\
\rowcolor{gray!5} Sonnet~4.6 & 10 & 18 & 1.8 & 1.5 & 0--4 \\
\bottomrule
\end{tabular}%
}
\caption{Observed archive-producing reflection/routing events by backbone. Each event corresponds to one escalation snapshot generated after a Coding \& Validation gate failure survives bounded in-loop repair; bounded repair attempts themselves are excluded by design. \textit{Per task} and \textit{Median} summarize the per-task archive-producing event count.}
\label{tab:reflection_envelope_main}
\end{table}

\paragraph{Routing event counts (Table~\ref{tab:telemetry_routing_count}).}
Each routing-recovery event creates a timestamped archived snapshot of the VSA. Counts therefore equal the number of times the pipeline left Coding \& Validation to redesign or reset within a task. Each (backbone, task) cell uses the final working directory of that task only.

\begin{table}[t]
\centering
\small
\renewcommand{\arraystretch}{1.15}
\resizebox{\columnwidth}{!}{%
\begin{tabular}{@{}l rrr@{}}
\toprule
\rowcolor{blue!8}
\textbf{Task(MLR-Bench)} & \textbf{Sonnet 4.5} & \textbf{Opus 4.5} & \textbf{Sonnet 4.6} \\
\midrule
\rowcolor{gray!5} bi\_align       & 2 & 3 & 2 \\
buildingtrust   & 0 & 7 & 1 \\
\rowcolor{gray!5} data\_problems  & 3 & 3 & 0 \\
dl4c            & 8 & 4 & 1 \\
\rowcolor{gray!5} mldpr           & 3 & 6 & 1 \\
question        & 7 & 11 & 4 \\
\rowcolor{gray!5} scope           & 0 & 3 & 2 \\
scsl            & 4 & 3 & 3 \\
\rowcolor{gray!5} verifai         & 1 & 0 & 3 \\
wsl             & 0 & 0 & 1 \\
\midrule
\textbf{Total}  & \textbf{28} & \textbf{40} & \textbf{18} \\
\bottomrule
\end{tabular}%
}
\caption{Routing event count per MLR-Bench e2e benchmark task and backbone. The 10 tasks cover selected MLR-Bench research topics, from alignment and trust to data, code, uncertainty, optimization, verification, and neural-network weights. Counts are computed from archived routing-recovery snapshots within each task's working directory.}
\label{tab:telemetry_routing_count}
\end{table}

\paragraph{Routing levels (Table~\ref{tab:telemetry_routing_levels}).}
Each archive snapshot is classified into one of the three Table~\ref{tab:recovery} routing levels using a cascade of signals (\texttt{episode.routing\allowbreak\_decision} $\rightarrow$ \texttt{termination\allowbreak\_trigger} + \texttt{routing\allowbreak\_reason} keywords $\rightarrow$ \texttt{failures\allowbreak\_by\_type} counts $\rightarrow$ \texttt{history} fallback). Snapshots without any parseable termination signal are reported as \textit{unclassified} and not discarded.

\begin{table}[t]
\centering
\small
\renewcommand{\arraystretch}{1.15}
\resizebox{\columnwidth}{!}{%
\begin{tabular}{@{}l rrrr@{}}
\toprule
\rowcolor{blue!8}
\textbf{Backbone} & \textbf{redesign} & \textbf{reset} & \textbf{unclass.} & \textbf{Total} \\
\midrule
\rowcolor{gray!5} Sonnet 4.5 & 3 & 25 & 0 & 28 \\
Opus 4.5             & 1 & 39 & 0 & 40 \\
\rowcolor{gray!5} Sonnet 4.6 & 1 & 17 & 0 & 18 \\
\midrule
\textbf{Total}       & \textbf{5} & \textbf{81} & \textbf{0} & \textbf{86} \\
\bottomrule
\end{tabular}%
}
\caption{Routing events by classified level \emph{among archive-producing events}, computed over the same final-retry-only working directories as Table~\ref{tab:telemetry_routing_count} (row totals match cell-wise). Bounded repair runs frequently in-loop within Coding \& Validation and produces no archive snapshot by design; this distribution therefore describes escalated failures only.}
\label{tab:telemetry_routing_levels}
\end{table}

Tables~\ref{tab:telemetry_routing_count}--\ref{tab:telemetry_routing_levels} together show 86 archive-producing reflection/routing events across the reported final-directory cells: 28 for Sonnet~4.5, 40 for Opus~4.5, and 18 for Sonnet~4.6. The compact event-count summary appears in Table~\ref{tab:reflection_envelope_main}.

\paragraph{Reflective Memory record format.}
Each routing event writes a structured record to Serena under the file pattern \texttt{failure\_\allowbreak\{hypothesis\_id\}\_\allowbreak run\{n\}.md} (or \texttt{pivot\_\allowbreak\{old\}\_\allowbreak\{new\}.md} for redesign-induced supersession). Subsequent Hypothesis Debate discussions consult these records before generating new candidates, using refuted approaches as constraints and partially validated elements as reusable evidence. Figure~\ref{fig:reflective_memory} shows an anonymized record drawn from one of the 30 benchmark runs.

\begin{figure}[t]
\begin{lstlisting}[basicstyle=\ttfamily\scriptsize,columns=fullflexible]
# Phase 4 Failure Record: h-e2 (Run 2)

Date: 2026-02-14T03:21:08
Hypothesis: h-e2
Final Status: FAIL
Failure Type: HYPOTHESIS_ISSUE

## Performance Gap
| Metric  | Ours | Baseline | Gap     |
|---------|------|----------|---------|
| WGA (%) | 71.3 | 78.6     | -7.3    |

## Root Causes
- CV-ratio threshold collapses on
  imbalanced minority groups
- Predicted monotonic separation does
  not hold under label noise > 5%

## Lessons Learned
- Threshold-based separation needs a
  robustness term for noisy labels
- Worst-group accuracy regresses when
  group size < 200

## Routing Decision
redesign -> Phase 2A (mechanism
revision: replace fixed threshold
with adaptive quantile)
\end{lstlisting}
\caption{Anonymized Reflective Memory record (Coding \& Validation failure, redesign route). Implementation labels are shown as recorded in the raw memory dump; the corresponding prose stage names are Coding \& Validation and Hypothesis Debate. The identifier \texttt{h-e\#} indexes sub-hypotheses within a parent hypothesis, and ``Run~2'' marks the second routing-recovery snapshot. The unedited Reflective Memory dumps that each benchmark run produced ship with the released repository.}
\label{fig:reflective_memory}
\end{figure}


\section{Supplementary Interpretation Details}\label{sec:benchmark_diag}

This appendix clarifies supplementary context behind the main results: tool-access differences, observed recovery-routing distribution and the 201-task upstream Idea/Proposal analysis. The automated diagnostic based on MLR-Bench's hallucination taxonomy is reported separately in Appendix~\ref{sec:hallucination_diagnostic}, and order-bias-controlled pairwise preferences remain in the main text as Table~\ref{tab:pairwise} because they are part of the primary result.

\subsection{Statistical Significance of the End-to-End Comparison}\label{sec:significance_test}
The unit of analysis is the matched (backbone, task) cell. Each cell score is the mean of the four LLM judges' Overall scores, and because the three systems share the same 30 cells, task and backbone variation cancels within pairs. We test the mean Overall difference with a paired sign-flip permutation test using 100{,}000 random sign assignments (two-sided). Table~\ref{tab:overall_significance} summarizes the results: YouRA leads MLR-Agent by $+0.85$ ($p{=}0.006$) and AI~Scientist~V2 by $+0.64$ ($p{=}0.044$), with both comparisons still significant under Holm correction (adjusted $p{=}0.012$ and $0.044$). The per-backbone mean differences are positive across all three backbones for both comparisons, so no single backbone drives the result. Cells sharing a backbone are not fully independent; the permutation runs at the cell level, not on individual judge verdicts, which avoids treating the 120 judge scores as independent samples.

\begin{table}[t]
\centering
\small

\caption{Paired significance tests for end-to-end Overall comparisons. The unit is the 30 matched (backbone, task) cells, where each cell averages four judges' Overall scores. Permutation $p$ is two-sided from 100{,}000 sign-flip assignments. W/L/T reports cell-level wins, losses, and ties for YouRA. Sign $p$ is the exact two-sided sign-test $p$ after excluding ties. Backbone deltas are reported in Sonnet~4.5 / Opus~4.5 / Sonnet~4.6 order.}
\label{tab:overall_significance}
\scriptsize
\renewcommand{\arraystretch}{2}
\resizebox{\columnwidth}{!}{%
\begin{tabular}{@{}lcccccc@{}}
\toprule
\rowcolor{blue!8}
\textbf{Comparison} & $\Delta$ \textbf{Overall} & \textbf{Perm.} $p$ & \textbf{Holm} $p$ & \textbf{W/L/T} & \textbf{Sign} $p$ & \textbf{Backbone deltas} \\
\midrule
YouRA $-$ MLR-Agent & $+0.85$ & 0.006 & 0.012 & 18/12/0 & 0.362 & $+1.10/+0.83/+0.63$ \\
YouRA $-$ AI Scientist V2 & $+0.64$ & 0.044 & 0.044 & 20/8/2 & 0.036 & $+0.58/+0.18/+1.18$ \\
\bottomrule
\end{tabular}
}
\end{table}

\subsection{201-Task Upstream Context}\label{sec:stagewise_context}
Table~\ref{tab:stagewise} compares Idea/Proposal-stage quality between YouRA and MLR-Agent on the full MLR-Bench task pool. Both systems were run with the Sonnet~4.5 and Opus~4.5 backbones, and a single Gemini~3.0~Pro judge scored both systems' outputs under the MLR-Bench stage rubrics. A Claude Opus~4.5 adapter at temperature 0 converted YouRA's hypothesis artifacts to the MLR-Bench idea/proposal format. Table~\ref{tab:stagewise_tests} reports task-paired sign-flip permutation tests. On Sonnet~4.5, MLR-Agent scores significantly higher on Overall at the Idea stage and on Clarity and Soundness at the Proposal stage; the Overall gap there does not survive correction. On Opus~4.5, Overall does not differ, with YouRA higher on Novelty and Feasibility and MLR-Agent higher on Soundness and Significance. This upstream comparison contrasts with the end-to-end paper-level comparison in Table~\ref{tab:e2e}, and it excludes AI~Scientist~V2 because that system was not run in this setting.

\begin{table*}[t]
\centering
\small
\renewcommand{\arraystretch}{1.25}
\resizebox{\textwidth}{!}{%
\begin{tabular}{lll ccccccc}
\toprule
\rowcolor{blue!8}
\textbf{Stage} & \textbf{System} & \textbf{Backbone LLM} &
\textbf{Consistency} & \textbf{Clarity} & \textbf{Novelty} &
\textbf{Soundness} & \textbf{Feasibility} &
\textbf{Significance} & \textbf{Overall} \\
\midrule

\rowcolor{gray!5}
 \multirow{4}{*}{Idea} & \multirow{2}{*}{YouRA} & Sonnet 4.5 & $9.49 \pm 1.65$ & $8.38 \pm 0.84$ & $7.32 \pm 0.88$ & -- & $6.75 \pm 1.44$ & $8.25 \pm 1.10$ & $7.90 \pm 1.37$ \\
  &  & Opus 4.5 & $9.91 \pm 0.60$ & $8.65 \pm 0.69$ & $\mathbf{7.49 \pm 0.76}$ & -- & $\mathbf{7.32 \pm 1.27}$ & $8.31 \pm 0.71$ & $8.40 \pm 0.83$ \\
  & \multirow{2}{*}{MLR-Agent} & Sonnet 4.5 & $\mathbf{9.97 \pm 0.30}$ & $\mathbf{8.62 \pm 0.67}$ & $7.19 \pm 0.71$ & -- & $6.76 \pm 1.03$ & $\mathbf{8.46 \pm 0.55}$ & $\mathbf{8.25 \pm 0.64}$ \\
  &  & Opus 4.5 & $9.98 \pm 0.17$ & $8.69 \pm 0.60$ & $7.24 \pm 0.68$ & -- & $6.98 \pm 1.07$ & $\mathbf{8.49 \pm 0.67}$ & $8.34 \pm 0.67$ \\
\midrule
\rowcolor{gray!5}
 \multirow{4}{*}{Proposal} & \multirow{2}{*}{YouRA} & Sonnet 4.5 & $9.74 \pm 0.87$ & $8.86 \pm 0.87$ & $7.89 \pm 0.85$ & $6.78 \pm 1.73$ & $7.00 \pm 1.55$ & $8.24 \pm 1.29$ & $7.79 \pm 1.66$ \\
  &  & Opus 4.5 & $\mathbf{9.92 \pm 0.28}$ & $9.09 \pm 0.68$ & $\mathbf{7.93 \pm 0.76}$ & $6.96 \pm 1.64$ & $\mathbf{7.58 \pm 1.46}$ & $8.17 \pm 1.04$ & $7.95 \pm 1.46$ \\
  & \multirow{2}{*}{MLR-Agent} & Sonnet 4.5 & $9.73 \pm 0.95$ & $\mathbf{9.12 \pm 0.81}$ & $7.72 \pm 1.25$ & $\mathbf{7.51 \pm 1.28}$ & $6.89 \pm 1.07$ & $8.48 \pm 0.94$ & $8.14 \pm 1.35$ \\
  &  & Opus 4.5 & $9.62 \pm 1.32$ & $9.11 \pm 0.56$ & $7.47 \pm 1.49$ & $\mathbf{7.42 \pm 1.40}$ & $7.14 \pm 1.21$ & $8.33 \pm 1.26$ & $8.06 \pm 1.56$ \\
\bottomrule
\end{tabular}%
}
\caption{Idea/Proposal-stage scores of YouRA and MLR-Agent on the MLR-Bench task pool. Both systems' outputs are scored by the same single Gemini~3.0~Pro judge under the MLR-Bench stage rubrics (1--10, higher better); YouRA's hypotheses are converted to the MLR-Bench idea/proposal format by a Claude Opus~4.5 adapter at temperature 0. ``--'' = metric not defined. Bold marks the higher mean where the paired difference is significant after Holm correction (Table~\ref{tab:stagewise_tests}).}
\label{tab:stagewise}
\end{table*}

\begin{table*}[t]
\centering
\caption{Paired tests for the Idea/Proposal-stage scores in Table~\ref{tab:stagewise} (YouRA minus MLR-Agent, matched by task). $\Delta$ Overall is the mean paired difference, Perm.\ $p$ is a two-sided sign-flip permutation $p$ (20{,}000 assignments), and Holm $p$ is corrected over the rubric dimensions within each stage--backbone block. The last two columns list the dimensions on which each system is significantly higher after correction.}
\label{tab:stagewise_tests}
\scriptsize
\setlength{\tabcolsep}{4pt}
\renewcommand{\arraystretch}{1.15}
\begin{tabular}{@{}llccc ll@{}}
\toprule
\rowcolor{blue!8}
\textbf{Stage} & \textbf{Backbone} & $\Delta$ \textbf{Overall} & \textbf{Perm.} $p$ & \textbf{Holm} $p$ & \textbf{YouRA higher} & \textbf{MLR-Agent higher} \\
\midrule
Idea & Sonnet 4.5 & $-0.35$ & 0.001 & 0.004 & -- & Cons., Clar., Sig., Overall \\
Idea & Opus 4.5 & $+0.08$ & 0.246 & 0.739 & Nov., Feas. & Sig. \\
Proposal & Sonnet 4.5 & $-0.36$ & 0.020 & 0.099 & -- & Clar., Sound. \\
Proposal & Opus 4.5 & $-0.11$ & 0.428 & 0.857 & Cons., Nov., Feas. & Sound. \\
\bottomrule
\end{tabular}
\end{table*}


\section{Automated Hallucination Diagnostic}\label{sec:hallucination_diagnostic}

\paragraph{Protocol.}
MLR-Bench's hallucination analysis defines four fact-based failure types (nonexistent or incorrect citations, hallucinated methodology, mathematical errors, and faked experimental results) and applies human verification after automated LLM detection~\citep{chen2025mlrbench}. We apply the automated detection stage to the generated papers across all three matched backbones, updating the original Gemini/Claude two-judge setup to a four-judge panel: Grok~4.3, GPT~5.4, Gemini~3.1~Pro, and Claude Opus~4.6. We report the results as judge-union and judge-intersection bounds, and validate the automatically flagged instances with three human annotators and an independent auditor (Table~\ref{tab:human_validation}; \S\ref{sec:human}).

\begin{figure*}[t]
\centering
\includegraphics[width=\textwidth]{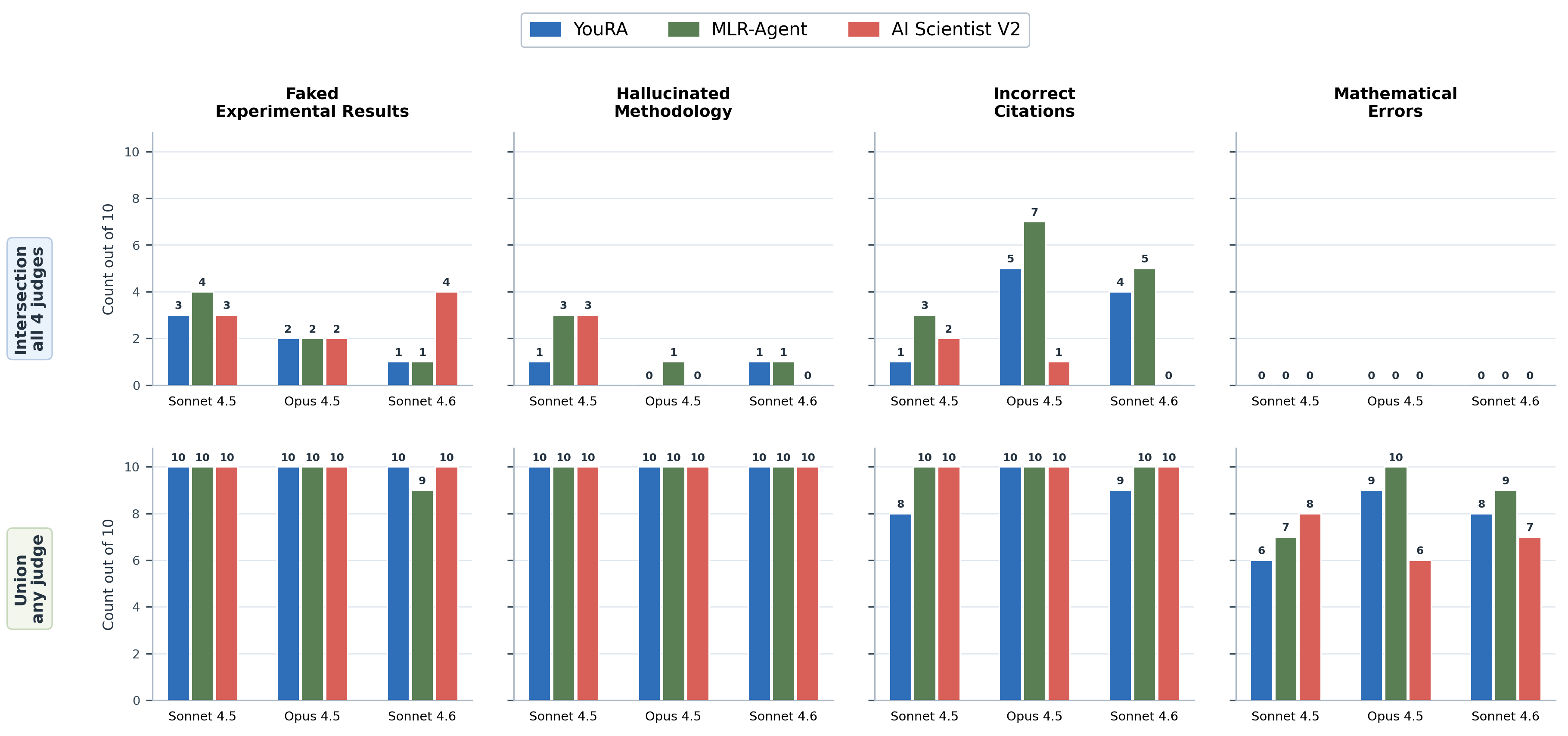}
\caption{Automated MLR-Bench hallucination diagnostic across matched backbones. The top row reports all-judge intersection counts out of ten papers, requiring all four judges to identify the hallucination type, while the bottom row reports any-judge union counts, requiring at least one judge to identify it. The four category columns denote nonexistent/incorrect citations (Cite), hallucinated methodology (Method), mathematical errors (Math), and faked experimental results (Faked). The diagnostic adapts MLR-Bench's automated hallucination categories but does not include the human verification step used in the original MLR-Bench hallucination analysis.}
\label{fig:hallucination_taxonomy_bounds}
\end{figure*}

\paragraph{Union/intersection semantics.}
Union counts capture any judge-flagged case, while intersection counts capture only cases flagged by all four judges. Figure~\ref{fig:hallucination_taxonomy_bounds} shows intersection counts in the top row and union counts in the bottom row; bar labels give the number of flagged papers out of ten. The gap between the two values reflects judge disagreement, and we report both instead of reducing the diagnostic to a single hallucination score.

\paragraph{Interpretive takeaway.}
Union counts are high for most rows, so we report the diagnostic as an automated profile of hallucination flags, and keep union and intersection counts separate; the profile is not used as a standalone ranking metric. On the all-judge-agreed intersection flags, YouRA shows a favorable profile on execution-evidence categories such as hallucinated methodology and faked experimental results; we report these as hallucination-flag counts; the human validation in \S\ref{sec:human} (Table~\ref{tab:human_validation}) estimates the precision of these automated flags at 75.2\% overall. Citation flags vary by backbone and are treated as a secondary signal. This pattern complements the Soundness/Overall gains in Table~\ref{tab:e2e} and the real-data provenance contrast in Figure~\ref{fig:vanilla_data_type} with concrete failure-type and data-provenance views.

\section{Data Provenance Analysis}\label{sec:provenance}

\paragraph{Protocol.}
Each generated paper is classified at the paper level by the kind of data its experiments ran on. An automated analyzer receives the paper and the root of its experiment folder, explores the folder without assuming its layout, and inspects the experiment code and result files to check that execution actually occurred, then assigns one of three labels: \emph{Real} (experiments run on real datasets, e.g., loaded through \texttt{datasets.load\_dataset} or \texttt{torchvision.datasets}), \emph{Synthetic} (experiments run on artificially generated data, e.g., random draws or synthetic-dataset generators), or \emph{Fabricated} (reported values without runnable experiment code, e.g., hardcoded result arrays). Every label is stored with the file-level evidence it rests on. Two analyzers built on distinct model families, Claude Code (Opus~4.6) and Codex (GPT-5.4), run the same protocol independently, so the classification does not hinge on a single model family; they agree on 85 of the 90 (system, backbone, task) cells. The counts reported in \S\ref{sec:data_provenance} use a conservative, maximum-synthetic reading: a run counts as real-data-based only when both analyzers label it \emph{Real}, so a \emph{Synthetic} or \emph{Fabricated} label from either analyzer counts against the run. Table~\ref{tab:provenance_backbone} breaks the counts down by backbone and analyzer.

\begin{table}[H]
\centering
\caption{Real-data classifications per system and backbone (out of ten tasks): the conservative both-analyzer count used in the main text, and each analyzer separately.}
\label{tab:provenance_backbone}
\footnotesize
\scriptsize
\setlength{\tabcolsep}{4pt}
\renewcommand{\arraystretch}{1.15}
\begin{tabular}{@{}llccc ll@{}}
\toprule
\rowcolor{blue!8}
\textbf{System} & \textbf{Backbone} & \textbf{Both analyzers} & \textbf{Opus 4.6} & \textbf{GPT-5.4} \\
\midrule
\multirow{3}{*}{YouRA} & Sonnet 4.5 & 8/10 & 8/10 & 9/10 \\
 & Opus 4.5 & 9/10 & 9/10 & 9/10 \\
 & Sonnet 4.6 & 10/10 & 10/10 & 10/10 \\
\midrule
\multirow{3}{*}{AI Scientist V2} & Sonnet 4.5 & 6/10 & 6/10 & 6/10 \\
 & Opus 4.5 & 8/10 & 9/10 & 8/10 \\
 & Sonnet 4.6 & 3/10 & 3/10 & 4/10 \\
\midrule
\multirow{3}{*}{MLR-Agent} & Sonnet 4.5 & 1/10 & 2/10 & 2/10 \\
 & Opus 4.5 & 4/10 & 4/10 & 4/10 \\
 & Sonnet 4.6 & 6/10 & 6/10 & 6/10 \\
\bottomrule
\end{tabular}
\end{table}

\paragraph{Counts by analyzer.}
Each analyzer taken alone gives the same ordering: 27/30, 12/30, and 18/30 real-data runs for YouRA, MLR-Agent, and AI~Scientist~V2 under the Opus~4.6 analyzer, and 28/30, 12/30, and 18/30 under GPT-5.4. All non-real cells are Synthetic except one MLR-Agent case labeled Fabricated. The analyzers assign labels automatically and record per-case evidence so that each classification can be inspected.

\section{Component Ablations}\label{sec:ablation}

This appendix interprets the ablation rows in main Table~\ref{tab:ablation} and reports their statistical tests. All four component ablations use Sonnet~4.5, Opus~4.5, and Sonnet~4.6. Each one removes a single component from full YouRA while leaving the rest of the workflow in place: \emph{w/o MCP} removes the MCP tool layer, \emph{w/o Reflection} disables reflection-guided redesign and reset, \emph{w/o VSA} removes the persistent state record that the model reads and updates, and \emph{w/o Independent Controller} removes the separate high-level controller. The analysis unit is the matched (backbone, task) cell. Each cell score averages the four judges' Overall scores, so every component ablation has 30 matched cells (three backbones $\times$ ten tasks).

\paragraph{Ablation significance.}
Table~\ref{tab:ablation_removal_significance} compares full YouRA with each ablated configuration over the same 30 cells. Removing any of the four components lowers \emph{Overall}, and all four differences stay significant after Holm correction under paired sign-flip permutation tests. These tests do not rank the components.

\begin{table}[H]
\centering
\caption{Paired significance tests for component removals. The unit is the 30 matched (backbone, task) cells, where each cell averages four judges' Overall scores. $\Delta$ Overall is ablated minus full YouRA, so negative values indicate drops under ablation. Permutation $p$ is two-sided from 100{,}000 sign-flip assignments. W/L/T reports cells where full YouRA is higher, lower, or tied. Backbone drops are reported in Sonnet~4.5 / Opus~4.5 / Sonnet~4.6 order.}
\label{tab:ablation_removal_significance}
\scriptsize
\setlength{\tabcolsep}{3.2pt}
\renewcommand{\arraystretch}{1.16}
\resizebox{\columnwidth}{!}{%
\begin{tabular}{@{}lcccccc@{}}
\toprule
\rowcolor{blue!8}
\textbf{Removed component} & $\Delta$ \textbf{Overall} & \textbf{Perm.} $p$ & \textbf{Holm} $p$ & \textbf{W/L/T} & \textbf{Sign} $p$ & \textbf{Backbone drops} \\
\midrule
MCP & $-1.13$ & $<0.001$ & $<0.001$ & 20/5/5 & 0.004 & $-1.38/-1.40/-0.63$ \\
Reflection & $-0.87$ & 0.007 & 0.007 & 17/11/2 & 0.345 & $-1.13/-0.85/-0.63$ \\
VSA & $-1.13$ & 0.002 & 0.004 & 22/8/0 & 0.016 & $-1.32/-1.12/-0.95$ \\
Independent Controller & $-1.12$ & 0.001 & 0.003 & 22/6/2 & 0.004 & $-1.82/-1.20/-0.35$ \\
\bottomrule
\end{tabular}
}
\end{table}

Table~\ref{tab:ablation_drop_contrasts} compares these drop sizes directly across removals. Every pairwise contrast has Holm-adjusted permutation $p{=}1.000$, so this sample cannot support a reliable ranking of components by drop size. The two analyses show that each component contributes while leaving their relative importance open.

\begin{table}[H]
\centering
\caption{Paired contrasts between component drop sizes. Each row tests the difference between two removal drops over the same 30 matched cells. Positive values mean that the first removal has the larger drop. W/L/T reports cells where the first drop is larger, smaller, or tied.}
\label{tab:ablation_drop_contrasts}
\scriptsize
\setlength{\tabcolsep}{4pt}
\renewcommand{\arraystretch}{1.15}
\begin{tabular}{@{}lcccc@{}}
\toprule
\rowcolor{blue!8}
\textbf{Drop-size contrast} & $\Delta$ \textbf{drop} & \textbf{Perm.} $p$ & \textbf{Holm} $p$ & \textbf{W/L/T} \\
\midrule
MCP $-$ Reflection & $+0.27$ & 0.169 & 1.000 & 17/9/4 \\
MCP $-$ VSA & $+0.00$ & 1.000 & 1.000 & 12/16/2 \\
MCP $-$ Independent Controller & $+0.01$ & 1.000 & 1.000 & 12/15/3 \\
Reflection $-$ VSA & $-0.27$ & 0.285 & 1.000 & 11/18/1 \\
Reflection $-$ Independent Controller & $-0.26$ & 0.252 & 1.000 & 11/16/3 \\
VSA $-$ Independent Controller & $+0.01$ & 1.000 & 1.000 & 17/9/4 \\
\bottomrule
\end{tabular}
\end{table}

\paragraph{Comparison with the baselines under ablation.}
On Sonnet~4.5, all four ablated configurations score below both MLR-Agent and AI~Scientist~V2 on Overall (Tables~\ref{tab:e2e} and~\ref{tab:ablation}). The same holds on Opus~4.5, although the w/o Reflection configuration is only marginally below MLR-Agent. On Sonnet~4.6, the ablated configurations all remain above AI~Scientist~V2; only w/o VSA falls below MLR-Agent, while w/o MCP and w/o Reflection tie it and w/o Independent Controller stays above. Either core-state removal (w/o VSA or w/o Independent Controller) drops YouRA below both baselines by clear margins on two of the three backbones (Sonnet~4.5 and Opus~4.5).

\paragraph{Backbone dependence.}
Across all four ablations, the drops are largest on Sonnet~4.5 and Opus~4.5 and smallest on Sonnet~4.6 (backbone-drop columns of Table~\ref{tab:ablation_removal_significance}), so a stronger execution backbone partially compensates for a missing component. The compensation is incomplete. The full configuration is still the highest-scoring YouRA variant on every backbone, and even the smallest drop (w/o Independent Controller on Sonnet~4.6) leaves the ablated configuration below the full one.

\paragraph{Failure profiles under core-component ablation.}
Judge justifications for the two Sonnet~4.5 core-component ablations show distinct failure signatures aligned with each component's role. Under w/o VSA, recurring flags include numbers that disagree across sections and tables, code implementing different experiments than the paper describes, claims contradicting execution logs, and selective reporting of favorable runs, all consistent with losing the single persistent record connecting hypotheses, gates, and evidence links. Under w/o Independent Controller, judges repeatedly flag failed or crashed runs whose outputs are still reported as completed findings, validation gates relaxed post hoc instead of escalated, and pipelines that shrink far below their plans without re-attempt, consistent with losing independent stop/continue judgment and recovery escalation. In both configurations Soundness is the most damaged dimension while Clarity stays high, matching the fluent-but-unsupported failure mode that motivates YouRA's design (\S\ref{sec:intro}).


\paragraph{Note on Clarity $\Delta$ under ablation.}
On Sonnet~4.5, the no-MCP configuration leaves Clarity nearly unchanged ($-0.05$); the no-reflection configuration shows a small Clarity increase ($+0.075$). On Sonnet~4.6, both the MCP and reflection ablations slightly reduce Clarity and increase Novelty. The Sonnet~4.5 VSA and controller ablations follow the same pattern: Clarity decreases only slightly ($-0.325$ and $-0.275$) while Soundness and Significance drop sharply. Across all configurations, the Overall drops are driven primarily by Soundness and Significance, the dimensions in which VSA-conditioned and reflection-informed reasoning is most heavily exercised; surface-level clarity contributes little.

\end{document}